\documentclass[11pt]{article}
\usepackage[preprint]{acl}
\usepackage{times}
\usepackage{latexsym}
\usepackage[T1]{fontenc}
\usepackage[utf8]{inputenc}
\usepackage{microtype}
\usepackage{inconsolata}

\usepackage{amsmath,amssymb}
\newtheorem{proposition}{Proposition}
\newtheorem{corollary}{Corollary}
\usepackage{booktabs}
\usepackage{multirow}
\usepackage{graphicx}
\usepackage{xcolor}
\usepackage{xspace}
\usepackage{algorithm}
\usepackage{algorithmic}
\usepackage{subcaption}
\usepackage{enumitem}
\usepackage{colortbl}
\usepackage{arydshln}   
\usepackage{framed}
\definecolor{PromptFrame}{RGB}{210,227,232}
\definecolor{PromptAccent}{RGB}{33,88,102}
\colorlet{TFFrameColor}{PromptFrame}
\colorlet{TFTitleColor}{PromptAccent}
\newlength{\PromptTitleWidth}

\newcommand{\PromptFrameBlock}[2]{%
  \begingroup
  \fboxsep=12pt\fboxrule=0.8pt
  \TitleBarFrame{%
    \parbox{\PromptTitleWidth}{%
      \normalfont\rmfamily\fontsize{11}{14}\selectfont\bfseries
      \strut #1\strut}}{#2}%
  \endgroup
}

\newenvironment{promptbox}[1]{%
  \par\begingroup
  \nolinenumbers
  \color{black}%
  \setlength{\PromptTitleWidth}{\dimexpr\linewidth-25.6pt\relax}%
  \setlength{\OuterFrameSep}{12pt}%
  \def\FrameCommand##1{\PromptFrameBlock{#1}{##1}}%
  \def\FirstFrameCommand##1{\PromptFrameBlock{#1}{##1}}%
  \def\MidFrameCommand##1{\PromptFrameBlock{#1\quad\normalfont\itshape(continued)}{##1}}%
  \let\LastFrameCommand\MidFrameCommand
  \MakeFramed{\advance\hsize-25.6pt\FrameRestore}%
  \def\small{\fontsize{10}{13}\selectfont}%
  \small\ttfamily\raggedright
  \setlength{\parindent}{0pt}%
  \setlength{\parskip}{5pt}%
  \hyphenpenalty=10000\exhyphenpenalty=10000
  \emergencystretch=2em
}{%
  \par\endMakeFramed\endgroup\color{black}%
}

\newcommand{\promptheading}[1]{{\color{PromptAccent}\bfseries #1}}
\newcommand{\promptlabel}[1]{{\bfseries #1}}
\newcommand{\promptdash}{{\normalfont\textendash}}

\definecolor{ourstint}{RGB}{214,236,249}
\definecolor{headgray}{gray}{0.90}
\definecolor{gaingreen}{RGB}{0,118,58}
\definecolor{Best}{HTML}{F5B7BF}
\definecolor{Secondbest}{HTML}{D5DEFF}
\definecolor{oursrow}{gray}{0.90}
\definecolor{bandblue}{HTML}{8080FF}
\newcommand{\gain}[1]{\textcolor{gaingreen}{\textbf{#1}}}

\newcommand{\method}{\textsc{BudgetAPO}\xspace}
\newcommand{\tse}{\tau}  
\newcommand{\ncand}{k}   
\newcommand{\bfirst}[1]{\cellcolor{Best}\textbf{#1}}
\newcommand{\usecond}[1]{\cellcolor{Secondbest}\underline{#1}}

\title{How Should a Prompt Optimizer Spend a Tight Budget?\\ \method with Noise-Adaptive Evaluation}

\author{
  Haoyue Liu\textsuperscript{1,2}\quad
  Zhichao Wang\textsuperscript{1}\quad
  Huanyu Yan\textsuperscript{1}\quad
  Xiaoying Tang\textsuperscript{1,2,3,*}\\[6pt]
  {\normalfont\small
  \textsuperscript{1}School of Science and Engineering, The Chinese University of Hong Kong, Shenzhen}\\
  {\normalfont\small
  \textsuperscript{2}Shenzhen Future Network of Intelligence Institute (FNii-Shenzhen)}\\
  {\normalfont\small
  \textsuperscript{3}Guangdong Provincial Key Laboratory of Future Networks of Intelligence, CUHK(SZ)}\\[3pt]
  {\normalfont\small
  \textsuperscript{*}Corresponding author.}
}

\begin{document}
\maketitle

\begin{abstract}
Automatic prompt optimization (APO) has been widely employed to adapt large language models without updating their weights, yielding promising results.
However, existing methods such as GEPA and OPRO assume hundreds to thousands of subject-model calls, far more than is practical behind paid, rate-limited APIs.
Under tight budgets they fail in two ways: multi-stage pipelines can exhaust the budget and return the seed prompt unchanged, while single-stage methods compare candidates on fixed-size minibatches, regardless of each task's noise.
As a remedy, we introduce \method, a single-stage optimizer for the tight-budget regime.
\method incorporates (1) a noise-adaptive rule that sizes the evaluation slice to each task's noise, measured by a short probe; (2) a fixed slice that turns every accept/reject decision into a paired comparison; and (3) a reflective operator that rewrites reasoning strategy and output format jointly.
Extensive results across seven benchmarks and five subject models demonstrate that \method ranks first on every subject and beats every baseline under Holm-corrected paired tests, while returning the seed in 13\% of runs at 250 calls against 86\% for GEPA.
On GPT-OSS-20B, GEPA needs 5 times as many calls to match \method's 100-call score.
\end{abstract}

\section{Introduction}
\label{sec:intro}

The performance of large language models (LLMs) is notoriously sensitive to the choice of prompt~\citep{brown2020language,reynolds2021prompt,wei2022chain,sclar2024quantifying}.
To adapt a model to a task without touching its weights, \emph{automatic prompt optimization} (APO) has emerged as the standard technique: given a task, a seed prompt and a validation set, an optimizer LLM searches for the prompt that maximizes the subject model's score~\citep{zhou2022large,yang2024large,pryzant2023automatic,opsahl2024optimizing,agrawal2026gepa}, often matching or beating hand-written prompts~\citep{zhou2022large,yang2024large}.

\begin{figure}[t]
  \centering
  \includegraphics[width=\columnwidth]{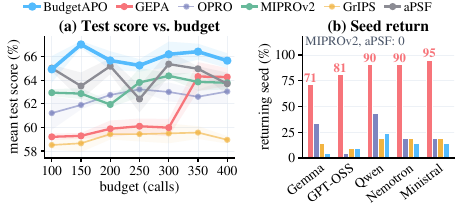}
  \caption{\textbf{(a)} Test score against budget, GPT-OSS-20B; bands: standard error over seeds.  \textbf{(b)} Seed-return rate at a 250-call cap, 105 runs per method.}
  \label{fig:teaser}
\end{figure}

Despite its success, APO has an unavoidable drawback: existing methods assume a generous evaluation budget.
GEPA~\citep{agrawal2026gepa} uses a multi-stage evaluation pipeline whose initial validation pass can consume several hundred subject-model calls, and OPRO~\citep{yang2024large} and MIPROv2~\citep{opsahl2024optimizing} were evaluated with a thousand or more.
In practice, however, subject models sit behind paid or rate-limited APIs, and every prompt is re-optimized whenever the model is updated.
Under such budgets the prevailing designs fail in two distinct ways (Figure~\ref{fig:teaser}).
First, \textbf{seed return}: a multi-stage pipeline can exhaust the budget before it selects a better candidate and then returns the prompt it was given; GEPA does so on 86\% of its tight-budget runs.
Second, \textbf{evaluation noise}: single-stage methods score candidates on minibatches of a fixed size, whatever the task's noise; on SuperGPQA, a 10-row batch carries a standard error of about 15 points.
Either way, the budget is spent without a reliable improvement.

\begin{figure*}[t]
  \centering
  \includegraphics[width=\textwidth]{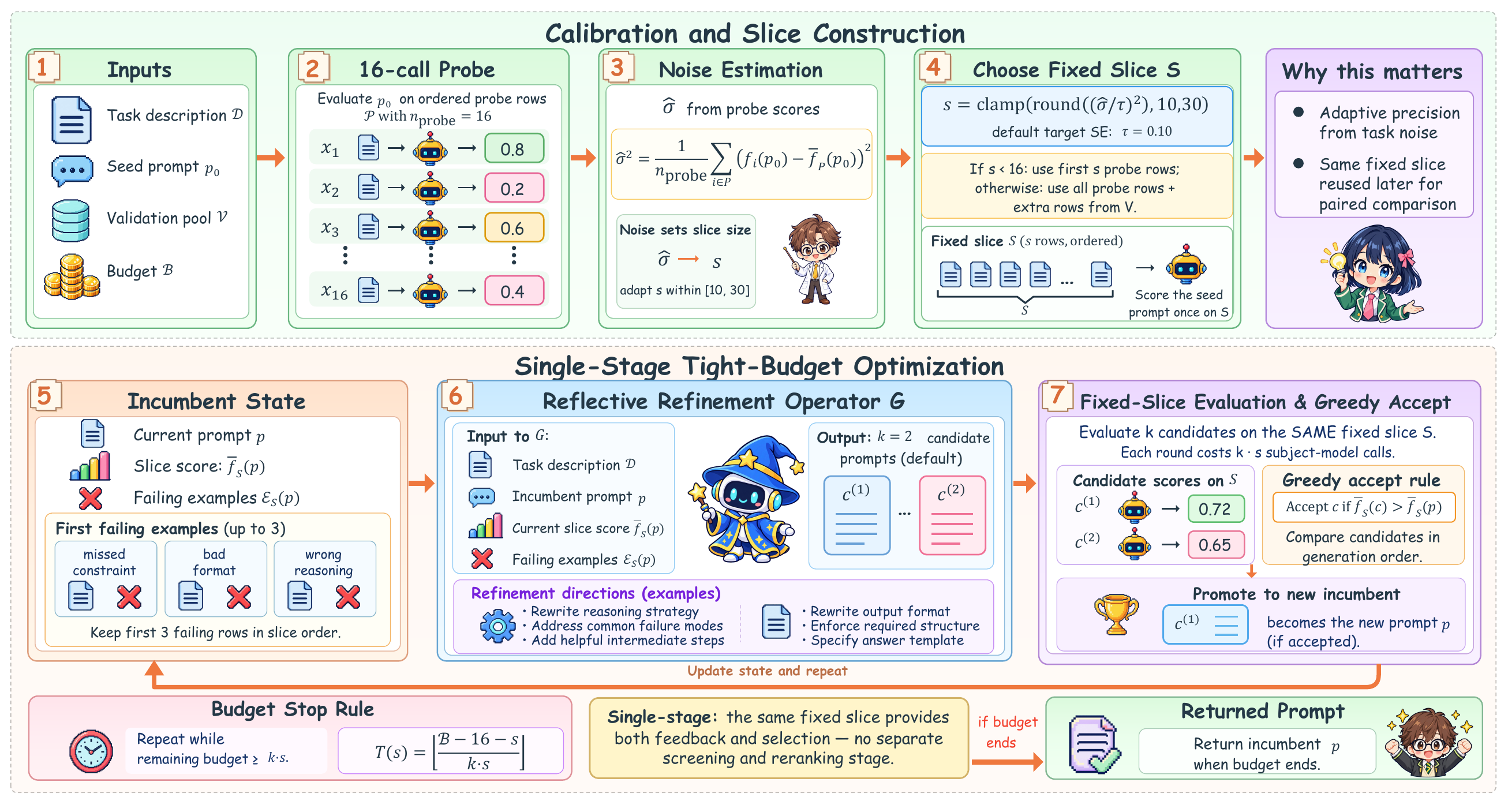}
  \caption{\textbf{One \method run.}
  Calibration spends 16 calls to estimate $\hat\sigma$,
  which fixes a slice of $s$ rows;
  each refinement round spends $k\,s$ calls
  (Equation~\ref{eq:steps}) and keeps a candidate
  only if it beats the incumbent
  (Equation~\ref{eq:accept}).}
  \label{fig:overview}
\end{figure*}

We therefore ask the following core question:
\begin{center}\itshape\interlinepenalty=10000
Can a tight-budget optimizer avoid seed return and size its evaluation to the task's noise?
\end{center}
Hence, we introduce \method, a single-stage optimizer with no separate validation pass, which addresses the aforementioned question through three components (Figure~\ref{fig:overview}): (1)~\emph{noise-adaptive slice sizing}, in which a 16-call probe measures the task's noise and a closed-form rule sizes the evaluation slice from it, thereby matching each comparison's precision to the task; (2)~a \emph{fixed evaluation slice} that scores every candidate for the rest of the run, turning each accept/reject decision into a paired comparison; and (3)~\emph{reflective refinement}, in which the optimizer rewrites reasoning strategy and output format jointly from a few failing rows.
We summarize our contributions as follows:
\begin{itemize}[leftmargin=1.2em,itemsep=2pt]
\item Introducing \method, the first APO method that automatically determines its evaluation size according to each task's measured noise before search, eliminating the need for per-task tuning with one set of defaults across all tasks and subject models.
\item Conducting extensive experiments against five baselines on seven benchmarks and five subject models (8B--120B, three seeds; 630 runs), showing that \method ranks first on every subject and significantly outperforms every baseline under Holm-corrected paired tests ($p<0.05$).
\item Demonstrating \method's efficiency under tight budgets (Figure~\ref{fig:teaser}): on GPT-OSS-20B, GEPA requires 5$\times$ as many calls to match \method's 100-call score, while at 250 calls it returns the seed in 86\% of runs compared with 13\% for \method.
\end{itemize}

\section{Related Work}
\label{sec:related}

Prompting shifts model behaviour without parameter updates, through few-shot examples~\citep{brown2020language} or elicited reasoning~\citep{wei2022chain,kojima2022large}, but the best prompt varies across models, tasks and inputs~\citep{liu2023pre}: formatting alone moves accuracy by several points~\citep{sclar2024quantifying}, and single-prompt evaluation can reverse a model ranking~\citep{mizrahi2024state}.  A line of work therefore automates the search.
APE~\citep{zhou2022large} selects among generated candidates; OPRO~\citep{yang2024large} shows the optimizer scored prompts and asks for a better one; GrIPS~\citep{prasad2023grips} edits the string; ProTeGi~\citep{pryzant2023automatic} propagates textual ``gradients''; PromptBreeder~\citep{fernando2023promptbreeder}, EvoPrompt~\citep{guo2024connecting} and EvoPrompting~\citep{chen2023evoprompting} evolve prompts; PromptAgent~\citep{wang2024promptagent} plans over an error tree.
MIPROv2~\citep{opsahl2024optimizing} extends DSPy~\citep{khattab2023dspy} with Bayesian optimization over instruction and demonstration configurations, and GEPA~\citep{agrawal2026gepa} screens a broad candidate pool on small batches and reranks survivors on larger ones; aPSF~\citep{liu2026adaptive} factorizes a prompt into semantic components and updates one at a time.
However, all of them were evaluated with hundreds to thousands of subject-model calls; our paper focuses on the regime below that.

Cheap evaluation is a cross-cutting concern: successive halving~\citep{li2018hyperband}, Bayesian search~\citep{bergstra2011algorithms} and meta-learned AutoML~\citep{feurer2022auto} early-stop unpromising hyper-parameter candidates, while \citet{polo2024tinybenchmarks} estimate benchmark scores from small subsets and \citet{vivek2024anchor} from a few anchor examples, trading 10--100$\times$ less cost for estimation noise; \citet{perlitz2024efficient} show how far a benchmark can be subsampled before its rankings move.
In APO the tradeoff is sharper, because subset noise interferes with the optimization signal itself.  CAPO~\citep{zehle2025capo} races candidates and TRIPLE~\citep{shi2024efficient} casts prompt selection as fixed-budget best-arm identification, both allocating evaluations per candidate during search.  \method instead calibrates one evaluation size to each task's noise before search with a closed-form rule, a design that also suits stacked-instruction tasks~\citep{anand2026instructionstackingcollapsebenchmark,liu2026bettercrosscomponentinterferencellm}.

\section{Method}
\label{sec:method}


A task supplies a validation pool $V$, a target evaluation distribution $\mathcal{R}$ over rows, and a scorer returning $f_i(p)\in[0,1]$ for prompt $p$ on row $i$.  With stochastic decoding, $f_i(p)$ scores one sampled reply.  For a slice $S\subseteq V$ of $s$ rows, write $\bar f_S(p)=\frac{1}{s}\sum_{i\in S} f_i(p)$.
One evaluation of a prompt on a validation row counts as one subject-model call, and $N_{\text{calls}}(\mathcal{A})$ denotes a run's call count.
Given a task description $D$, a seed prompt $p_0$ and a budget $B$, a run returns one prompt $\hat p_B = \mathcal{A}(D, p_0, V; B)$; the object of study is the optimizer:
\begin{equation}
  \max_{\mathcal{A}\,:\,N_{\text{calls}}(\mathcal{A})\,\le\, B}\;
  \mathbb{E}\big[f_I(\hat p_B)\big],\qquad I\sim\mathcal{R},
  \label{eq:obj}
\end{equation}
The expectation averages over a fresh target row $I$, optimization randomness, and decoding randomness; the run searches on $V$ and is scored on the test pool, whose distribution may differ, as in IFBench.
In this section, we introduce \method, an algorithm capable of sizing its evaluation to each task's noise within the budget of Equation~\ref{eq:obj}.  This is achieved through three crucial components (Figure~\ref{fig:overview}): (1)~noise-adaptive slice sizing and (2)~a fixed evaluation slice (\S\ref{sec:slice}), and (3)~a reflective refinement operator (\S\ref{sec:operator}), with greedy acceptance (\S\ref{sec:accept}); Algorithm~\ref{alg:main} summarizes the process.  Here \emph{single-stage} means that one slice supplies both refinement feedback and selection, with no separate screening and reranking stage.

\begin{figure}[t]
  \centering
  \includegraphics[width=\columnwidth]{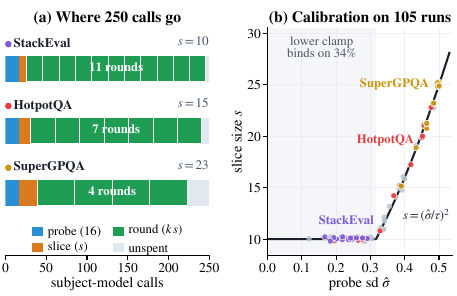}
  \caption{\textbf{(a)} How a 250-call cap divides at each benchmark's mean slice size (Equation~\ref{eq:steps}): a larger slice buys fewer rounds.  \textbf{(b)} Probe standard deviation $\hat\sigma$ against the chosen slice, one point per run; the tasks of (a) in color.}
  \label{fig:method}
\end{figure}

\subsection{Noise-Adaptive Slice Sizing}
\label{sec:slice}

In this section, we present noise-adaptive slice sizing, which eliminates the need for choosing a minibatch size per task.

\paragraph{Fixed minibatches and their limitations.}
Single-stage methods score candidates on minibatches of a fixed size $m$, whose standard error $\sigma/\sqrt{m}$ depends on the task's per-row standard deviation $\sigma$.  Two limitations follow: 1.~on a noisy task comparisons are imprecise (10 SuperGPQA rows, $\sigma\approx0.48$, give a standard error near 0.15); 2.~on a stable task extra rows buy no precision but cost search depth (Equation~\ref{eq:steps}).

\paragraph{Addressing the limitations by noise-adaptive sizing.}
To address them, \method measures the noise first and targets a standard error $\tse$:
\begin{equation}
  \mathrm{SE}\big[\bar f_S(p)\big] = \frac{\sigma}{\sqrt{s}},
  \qquad
  \mathrm{SE}=\tse \;\Rightarrow\; s=\Big(\frac{\sigma}{\tse}\Big)^{2},
  \label{eq:se}
\end{equation}
To illustrate, we first score the seed prompt $p_0$ on an ordered probe sample $P\subset V$ of $n_\text{probe}=16$ rows and read off the per-row spread,
\begin{equation}
  \hat{\sigma}^{2} \;=\; \frac{1}{n_\text{probe}}\sum_{i\in P}\big(f_i(p_0)-\bar f_P(p_0)\big)^{2},
  \label{eq:sigma}
\end{equation}
Then the slice size is set as
\begin{equation}
  s = \mathrm{clamp}\!\left(\Big\lfloor\Big(\frac{\hat{\sigma}}{\tse}\Big)^{\!2}\Big\rceil,\; s_{\min},\; s_{\max}\right),
  \label{eq:slice}
\end{equation}
where $\lfloor\cdot\rceil$ rounds to the nearest integer (ties to even), $\mathrm{clamp}(x,a,b)=\min\{b,\max\{a,x\}\}$, the default target is $\tse=0.10$, and $s_{\min}=10$, $s_{\max}=30$.
Finally, the slice $S$ is drawn once and \emph{fixed}: it takes the first $s$ probe rows when $s\le n_\text{probe}$, or all of $P$ topped up from $V\setminus P$ otherwise, and scores every candidate for the rest of the run.
In the archived logs (Figure~\ref{fig:method}b), StackEval's $\hat\sigma$ averages 0.23 ($s=s_{\min}=10$) and SuperGPQA's 0.48 (about 23 rows); the lower clamp binds in 34\% of 105 runs, and $s_{\max}$ never does at the default $\tse$, since $\hat\sigma\leq0.50$.
With $\ncand$ candidates per step and the seed re-scored on $S$ (\S\ref{sec:accept}), $B\geq n_\text{probe}+s$ permits
\begin{equation}
  T(s) \;=\; \left\lfloor \frac{B - n_\text{probe} - s}{\ncand \cdot s} \right\rfloor .
  \label{eq:steps}
\end{equation}
At $B=250$ and $\ncand=2$ this is 11 steps at $s=10$ and 4 at $s=25$: every row added to the slice is bought with search depth (Figure~\ref{fig:method}a).
Scoring a candidate and the incumbent on the same rows reduces the variance of their difference under positive same-row covariance (Appendix~\ref{app:theory}); reusing the slice is worth 1.4 points (\S\ref{sec:ablation}), and on held-out rows the prompt it selects beats its candidate-pool mean by a median of 1.7 points (Appendix~\ref{app:gengap}).

\paragraph{Remark 3.1 (Discussion on the plug-in estimate).}
The seed's spread stands in for that of candidates not yet written; Appendix~\ref{app:theory} treats the accept/reject decision, which turns on the variance of a \emph{difference}.

\subsection{Reflective Refinement Operator}
\label{sec:operator}

\paragraph{Existing feedback and its limitations.}
OPRO~\citep{yang2024large} shows the optimizer past prompts with their scores, and ProTeGi~\citep{pryzant2023automatic} derives textual ``gradients'' from errors.  Under a tight budget two limitations matter: 1.~a score or an aggregate does not say on what a prompt fails; 2.~a rewrite that improves reasoning can break the output format.

\paragraph{Addressing the limitations by reflective refinement.}
To address them, the operator shows raw failing rows and asks for strategy and format together.  At each step the optimizer LLM $G$ receives the task description $D$, the incumbent $p$ with its recorded slice score, and an ordered list of \emph{failing examples} $E_S(p)$, and returns $\ncand$ candidates in generation order (default $\ncand=2$):
\begin{equation}
\begin{aligned}
  (c^{(1)},\dots,c^{(\ncand)}) &= G\big(D,\,p,\,\bar f_S(p),\,E_S(p)\big),\\
  E_S(p) &= \big(\,i\in S : f_i(p)<1\,\big)_{[:3]},
\end{aligned}
  \label{eq:operator}
\end{equation}
where $E_S(p)$ lists the failing rows of $S$ in slice order and $[:3]$ keeps the first three, each shown with the request, the constraints it missed, and the start of the reply.
$G$ rewrites both \textbf{strategy}, how the subject model should reason about the task, and \textbf{format}, what the output should look like in structure, delimiters and length.
The template is in Appendix~\ref{app:template}.

Raw rows beat aggregates: replacing one with a per-family failure histogram costs 8.8 points across nine cells, and adding the histogram costs 5.1 on IFBench (Appendix~\ref{app:histogram}).

\subsection{Greedy Accept with Best Tracking}
\label{sec:accept}

In this section, we present the acceptance rule, which uses every rollout for both feedback and selection.
At step $t$ the operator proposes $\ncand$ candidates $c_t^{(1)},\dots,c_t^{(\ncand)}$, each scored on the fixed slice and compared to the incumbent in generation order:
\begin{equation}
  p \leftarrow c_t^{(i)} \quad\text{if}\quad \bar f_S\big(c_t^{(i)}\big) > \bar f_S(p),
  \label{eq:accept}
\end{equation}
for $i=1,\dots,\ncand$ in turn, so a later candidate is judged against an earlier one if that one was taken.
Writing $B_\text{used}$ for the calls already spent, the run ends when $B - B_\text{used}$ no longer covers a step of $\ncand\, s$ calls, and returns the incumbent.

Following GEPA~\citep{agrawal2026gepa}, the budget counts only \emph{subject-model calls} (scoring candidates on validation rows); optimizer-model calls and the final test-set evaluation are free for all methods (Appendix~\ref{app:setup}).


\section{Experiments}
\label{sec:exp}

\begin{table*}[t]
\centering
\footnotesize
\setlength{\tabcolsep}{4.4pt}
\caption{\textbf{Main result} (test score in \%, budget 250, seeds 0/1/2).  \colorbox{Best}{\textbf{Bold}} best, \colorbox{Secondbest}{\underline{underline}} second; $\pm$: standard deviation over the three seed-level means, in (b) over the 15 subject--seed scores.  $\Delta$: gap to the row's strongest baseline.  \emph{No opt.}: the seed prompt at seed 0.  \emph{Baseline spread}: best minus worst baseline.  Per cell: Table~\ref{tab:main_big}.}
\label{tab:main5}
\vspace{1pt}
{\footnotesize\textbf{(a) By subject model.}}\\[3pt]
\renewcommand{\arraystretch}{1.35}
\setlength{\tabcolsep}{9.4pt}
\begin{tabular}{l cccccc r}
\toprule
\rowcolor{headgray} \textbf{Subject model} & \textbf{\method} & \textbf{GEPA} & \textbf{OPRO} & \textbf{MIPROv2} & \textbf{GrIPS} & \textbf{aPSF} & $\Delta$ \\
\midrule
Gemma-4-31B & \bfirst{71.57}{\scriptsize$\pm$0.50} & 67.11 & 67.58 & 66.62 & 66.97 & \usecond{68.85} & \gain{+2.72} \\
GPT-OSS-20B & \bfirst{65.24}{\scriptsize$\pm$0.91} & 60.10 & 63.22 & \usecond{63.80} & 59.43 & 62.41 & \gain{+1.44} \\
Qwen-3.5-27B & \bfirst{67.40}{\scriptsize$\pm$2.13} & 66.11 & 65.78 & \usecond{66.16} & 65.56 & 64.18 & \gain{+1.24} \\
Nemotron-3-Super-120B~~ & \bfirst{61.43}{\scriptsize$\pm$0.22} & 58.51 & \usecond{60.43} & 58.80 & 58.93 & 59.61 & \gain{+1.00} \\
Ministral-8B & \bfirst{51.98}{\scriptsize$\pm$1.63} & 47.83 & \usecond{50.72} & 49.76 & 47.99 & \usecond{50.72} & \gain{+1.26} \\
\midrule
\textbf{Mean of 5} & \bfirst{63.52}{\scriptsize$\pm$0.96} & 59.93 & \usecond{61.54} & 61.03 & 59.78 & 61.15 & \gain{+1.98} \\
\bottomrule
\end{tabular}

\vspace{9pt}
{\footnotesize\textbf{(b) By benchmark, averaged over the five subjects.}}\\[3pt]
\footnotesize
\setlength{\tabcolsep}{4.6pt}
\renewcommand{\arraystretch}{1.35}
\begin{tabular}{lccccccc|c}
\toprule
\rowcolor{headgray} \textbf{Method} & \textbf{Hotpot} & \textbf{CollIE} & \textbf{IFBench} & \textbf{SuperGPQA} & \textbf{Stack} & \textbf{MBPP+} & \textbf{LB-IF} & \textbf{Mean} \\
\midrule
\emph{No opt.} & 61.50 & 52.80 & 44.80 & 42.60 & 71.37 & 76.53 & 63.26 & 58.98 \\
\rowcolor{bandblue!8} \multicolumn{9}{l}{\textcolor{bandblue}{\textit{Published baselines}}} \\
GEPA & 64.60{\scriptsize$\pm$12.91} & 54.02{\scriptsize$\pm$14.15} & 46.12{\scriptsize$\pm$9.05} & 42.37{\scriptsize$\pm$12.79} & 71.29{\scriptsize$\pm$6.65} & \bfirst{77.41}{\scriptsize$\pm$4.47} & 63.72{\scriptsize$\pm$6.22} & 59.93 \\
OPRO & 67.55{\scriptsize$\pm$11.29} & 57.30{\scriptsize$\pm$15.61} & 50.58{\scriptsize$\pm$11.16} & 44.13{\scriptsize$\pm$10.38} & \usecond{71.85}{\scriptsize$\pm$6.11} & 75.42{\scriptsize$\pm$4.16} & \usecond{63.97}{\scriptsize$\pm$5.46} & \usecond{61.54} \\
MIPROv2 & \usecond{69.79}{\scriptsize$\pm$12.83} & 57.60{\scriptsize$\pm$15.27} & 50.75{\scriptsize$\pm$12.05} & 42.43{\scriptsize$\pm$12.75} & 69.17{\scriptsize$\pm$7.36} & 74.47{\scriptsize$\pm$4.02} & 63.00{\scriptsize$\pm$7.66} & 61.03 \\
GrIPS & 66.15{\scriptsize$\pm$11.52} & 52.92{\scriptsize$\pm$13.63} & 47.15{\scriptsize$\pm$9.34} & 42.07{\scriptsize$\pm$12.63} & 70.92{\scriptsize$\pm$6.69} & 76.10{\scriptsize$\pm$4.86} & 63.13{\scriptsize$\pm$6.23} & 59.78 \\
aPSF & 69.25{\scriptsize$\pm$8.54} & \usecond{58.10}{\scriptsize$\pm$15.85} & \bfirst{52.85}{\scriptsize$\pm$11.04} & \usecond{44.18}{\scriptsize$\pm$10.73} & 69.55{\scriptsize$\pm$6.62} & 70.64{\scriptsize$\pm$7.76} & 63.49{\scriptsize$\pm$5.97} & 61.15 \\
\cdashline{1-9}
\rowcolor{oursrow} \method & \bfirst{73.29}{\scriptsize$\pm$8.85} & \bfirst{59.50}{\scriptsize$\pm$17.40} & \usecond{52.00}{\scriptsize$\pm$11.01} & \bfirst{45.33}{\scriptsize$\pm$11.54} & \bfirst{73.00}{\scriptsize$\pm$5.02} & \usecond{77.30}{\scriptsize$\pm$4.14} & \bfirst{64.23}{\scriptsize$\pm$6.11} & \bfirst{63.52} \\
\midrule
\emph{Baseline spread} & 5.19 & 5.18 & 6.73 & 2.11 & 2.68 & 6.77 & 0.97 & 1.76 \\
\bottomrule
\end{tabular}

\end{table*}

In this section, we carry out experiments to address the following questions:
\begin{itemize}[leftmargin=1.2em,itemsep=1pt]
\item \textbf{Q1}: Can \method beat existing APO methods at 250 calls on every subject model?  See \S\ref{sec:main}.
\item \textbf{Q2}: Is \method more call-efficient than existing methods when the budget is tight?  See \S\ref{sec:budget_results}.
\item \textbf{Q3}: Which components drive the gains, and is \method robust to its hyper-parameters?  See \S\ref{sec:ablation}.
\item \textbf{Q4}: Does the advantage hold under a different optimizer LLM?  See \S\ref{sec:optimizer}.
\end{itemize}
Additional results, including (1) per-cell tables, (2) the full budget curve and (3) case studies, are in Appendices~\ref{app:per_subject}--\ref{app:casestudy}.

\subsection{Setup}
\label{sec:setup}
To answer these questions, we evaluate six APO methods with five subject models on seven benchmarks, as detailed below.

\paragraph{Subject models.}
The main comparison uses five subject models spanning 8B--120B parameters and five families: Gemma-4-31B, Qwen-3.5-27B, GPT-OSS-20B, Nemotron-3-Super-120B and Ministral-8B (endpoints in Appendix~\ref{app:setup}).
Gemma-4-31B~\citep{team2026gemma} is our \emph{primary} subject, on which the ablations, hyper-parameter sweeps and optimizer swap run; the rewrite and seed-return counts pool all five subjects.
Within each subject, all methods share the decoding configuration and the same five-word seed, ``You are a helpful assistant.''  Temperature is 0.6 and top-$p$ is 0.95.  The default optimizer is GLM-5-FP8~\citep{zeng2026glm}; \S\ref{sec:optimizer} swaps in an alternative.

\paragraph{Benchmarks.}
Seven tasks span multi-hop QA, instruction following, graduate-level knowledge and code: HotpotQA~\citep{yang2018hotpotqa}, CollIE~\citep{yao2024collie}, IFBench~\citep{pyatkin2025generalizingverifiableinstructionfollowing}, SuperGPQA~\citep{du2026supergpqa}, a generated instruction-stacking task (labeled StackEval in our tables), MBPP+~\citep{liu2023your} and LiveBenchIF~\citep{white2025livebench}.  We report each task's normalized score on a percentage scale and average tasks with equal weight.

\paragraph{Baselines.}
We compare against five APO methods using their official implementations: \textbf{GEPA}~\citep{agrawal2026gepa}, \textbf{MIPROv2}~\citep{opsahl2024optimizing}, \textbf{OPRO}~\citep{yang2024large}, \textbf{GrIPS}~\citep{prasad2023grips}, and \textbf{aPSF}~\citep{liu2026adaptive}.

\paragraph{Evaluation protocol.}
We instantiate the tight-budget regime with a cap of $B=250$ subject-model calls per run; every method receives the same cap.  The returned prompt is scored on a held-out test pool shared by all methods, outside the budget.  Each (subject, method, benchmark) triple is reported over seeds 0, 1 and 2 as mean test score $\pm$ sample standard deviation; runs failing a coverage or budget guard were re-run (Appendix~\ref{app:setup}).
Significance is assessed by two-sided paired Wilcoxon signed-rank tests with Holm correction~\citep{holm1979simple,wilcoxon1945individual} over the $n=35$ (subject, benchmark) pairs, seeds averaged within a cell so each pair contributes one observation.  Effect sizes are Cohen's $d_z$~\citep{cohen2013statistical}: the mean paired difference divided by its sample standard deviation.

\subsection{\texorpdfstring{A1: \method Leads on Every Subject}{A1: BudgetAPO Leads on Every Subject}}
\label{sec:main}

\begin{figure*}[t]
  \centering
  \includegraphics[width=\textwidth]{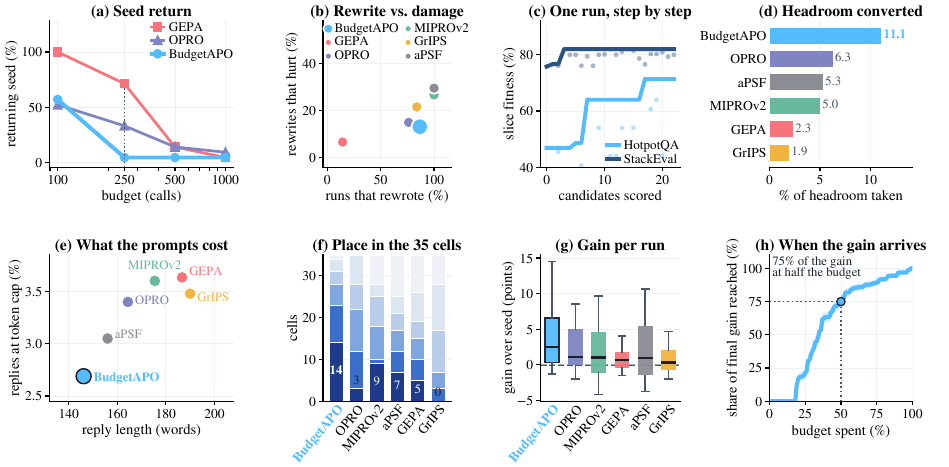}
  \caption{Analysis at a 250-call cap, except (a).  \textbf{(a)} Seed returns against budget, primary subject.  \textbf{(b)} Rewrite rate against the share of rewrites more than one point below the seed.  \textbf{(c)} Candidate scores and running best in one HotpotQA and one StackEval run.  \textbf{(d)} Gain as a share of the headroom to 100.  \textbf{(e)} Reply length and share truncated at the token cap.  \textbf{(f)} Place over the 35 cells, darkest = first.  \textbf{(g)} Gain over the seed per run; whiskers: 10th--90th percentiles.  \textbf{(h)} Share of the final slice gain reached against budget spent (91 runs); band: standard error.  (b), (d), (e), (g) pool 105 runs per method.}
  \label{fig:analysis}
\end{figure*}

In this section, we compare all six methods on the five subject models at budget 250 (Table~\ref{tab:main5}; per cell in Table~\ref{tab:main_big}).  We can observe that 1.~\method ranks first on all five, by $+1.00$ points on Nemotron-3-Super-120B up to $+2.72$ on Gemma-4-31B, and reaches 63.52\% on the five-subject mean against 61.54\% for the best baseline (OPRO), a gap of $+1.98$ points; 2.~the runner-up changes from subject to subject (aPSF, MIPROv2, OPRO) and each also finishes fourth or worse elsewhere, so no baseline is a safe pick before the deployment subject is known; 3.~across the 35 (subject, benchmark) cells, \method ranks first in 14, including two shared first places, and is in the top two in 23, against 12 for the next best method (Figure~\ref{fig:analysis}f).

\paragraph{The lead is significant against every baseline.}
Table~\ref{tab:sig} tests \method against each baseline separately over the 35 (subject, benchmark) cells: it wins 23--29 of them and clears the Holm-corrected threshold against all five, with effect sizes $d_z=0.47$--$0.82$.
Enlarging four of the primary subject's test sets to 731--1\,000 rows preserves \method's position on those tasks (Appendix~\ref{app:xl}).
On the primary subject, \method's seed-to-seed standard deviation is 0.50 points, level with GrIPS (0.48) and MIPROv2 (0.50) and below OPRO (0.91), GEPA (1.38) and aPSF (1.86); on the five-subject mean it is 0.96 (Appendix~\ref{app:variance}).

On the primary subject (first block of Table~\ref{tab:main_big}) \method is first on five of seven benchmarks, with the largest gain on HotpotQA ($+11.6$ over the next best), and lifts the seed prompt's 65.6\% to 71.6\% on the mean.
On HotpotQA the gap over GEPA is 20.0 points (74.3 vs.\ 54.3): GEPA returns the unchanged seed on all three seeds and \method on none.  All three of its HotpotQA runs use $s=10$, allowing 11 refinement steps by Equation~\ref{eq:steps}.
Averaging over subjects rather than benchmarks gives the complementary view in Table~\ref{tab:main5}(b): first on five of seven and second on the other two.  On the mean, \method's lead over the best baseline ($+1.98$) is larger than the \emph{Baseline spread}, the gap between the best and worst baselines (1.76; Appendix~\ref{app:spread}).

\paragraph{\method gains the most per run.}
\label{sec:rewrite}
Figure~\ref{fig:analysis}b pairs rewrite frequency (a rewrite is a returned prompt that differs from the seed) with the share of rewrites more than one point below the seed; the gain of a run is its test score minus the seed prompt's score in the same cell, the \emph{No optimization} entry of Table~\ref{tab:main_big} (on HotpotQA, from the seed-0 pool).
\method rewrites in 91 of 105 runs (87\%), 13\% of them more than a point below the seed; MIPROv2 and aPSF rewrite in all 105 (27\% and 30\%), GrIPS in 88 (22\%), OPRO in 80 (15\%), and GEPA in 15 (7\%, with a mean gain of $+0.95$ points over all runs).
Across all 105 runs, \method's mean gain is $+4.54$ points, against $+2.56$ for the best baseline; its median is $+2.5$ points against at most $+1.1$ (Figure~\ref{fig:analysis}g).  Dividing the summed gains by the summed headroom from the seed's score to 100 gives 11.1\% for \method against 1.9--6.3\% for the baselines (Figure~\ref{fig:analysis}d).

\begin{figure}[t]
  \centering
  \includegraphics[width=0.86\columnwidth]{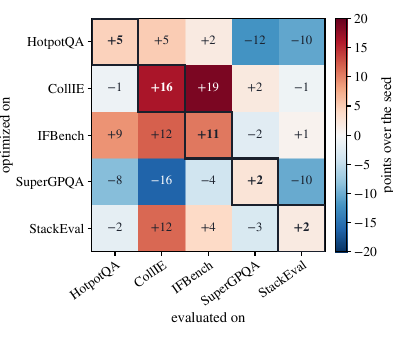}
  \caption{\textbf{Prompt transfer} on five of the seven tasks.  Change against the seed prompt when the prompt optimized on the row's task is scored on the column's test pool (GPT-OSS-20B, budget 250).  Boxed: the source task.}
  \label{fig:transfer}
\end{figure}

\begin{figure}[t]
  \centering
  \includegraphics[width=\columnwidth]{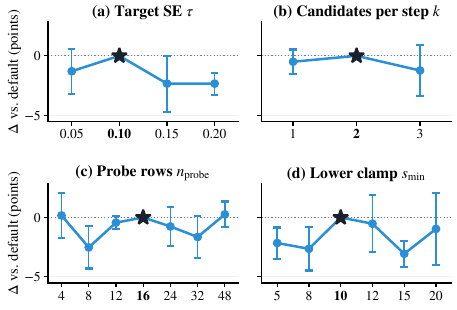}
  \caption{\textbf{Hyper-parameter sensitivity}, primary subject: change in the seven-benchmark mean against the default ($\star$), paired by seed; bars: seed standard deviation. Per-benchmark results are in Appendix~\ref{app:sens}.}
  \label{fig:sens}
\end{figure}

\paragraph{The prompts are task-specific and draw short replies.}
\method's returned prompts have a median length of 204 words, compared with 39 for MIPROv2 and 5 for GEPA.
On the five tasks of Figure~\ref{fig:transfer}, a prompt gains $+7.1$ points over the seed on its source task and $-0.3$ on the others, and the instruction-following tasks CollIE and IFBench share gains of $+19$ and $+12$: on average, the gain stays with the task a prompt was written for.
Over the 105 main runs, \method's prompts draw the shortest replies of any method, 146 words against 156--190, and the fewest truncated at the token cap, 2.7\% against 3.1--3.6\% (Figure~\ref{fig:analysis}e).

\subsection{A2: The Gain Is Largest Where the Budget Is Tightest}
\label{sec:budget_results}

In this section, we vary the budget from 100 to 400 calls in steps of 50 on GPT-OSS-20B for all six methods (three seeds, seven benchmarks, 882 runs; Figure~\ref{fig:teaser}a).  We can observe that 1.~\method is first at every sampled budget from 150 to 400, by 0.5--3.5 points over the strongest baseline: it scores 64.9\% at 100, peaks at 67.0\% at 150, and ranges from 65.2 to 66.4\% through 400; 2.~GEPA scores 59.2--60.1\% from 100 to 300, trailing by 5.1--7.7 points, and rises to 64.3\% at 350, once the cap fits its initial 300-row validation pass on five of the seven tasks; 3.~even so, GEPA needs 500 calls, 5 times as many, to match \method's 100-call score (65.1 against 64.9), MIPROv2 ties it only at 500, and OPRO and GrIPS stay below it through 500; 4.~\method's 150-call score of 67.0\% is the highest any method reaches at any budget, and averaged over the seven budgets \method scores 65.9\% against 64.3\% for the strongest baseline, aPSF, and 61.0\% for GEPA.

\begin{table}[t]
\centering
\footnotesize
\setlength{\tabcolsep}{3.7pt}
\caption{Ablations, primary subject (\%, budget 250); each column removes one component against the shaded \textbf{Full} entry.  $-$Adapt.: fixed $s{=}20$, probe still spent.  $-$Fixed: slice redrawn every step, incumbent re-scored on it.  $-$Strat.: plain ``raise the score'' request, no failing rows.  $-$Fail.: strategy and format kept, no failing rows.}
\label{tab:design}
\begin{tabular}{l|>{\columncolor{ourstint}}c|cccc}
\toprule
\rowcolor{headgray} \textbf{Benchmark} & \textbf{Full} & $-$Adapt. & $-$Fixed & $-$Strat. & $-$Fail. \\
\midrule
HotpotQA & 74.3 & 66.1 & 65.9 & 68.8 & 73.2 \\
CollIE & 74.2 & 70.8 & 72.7 & 69.8 & 70.0 \\
IFBench & 62.7 & 62.7 & 65.2 & 61.2 & 63.6 \\
SuperGPQA & 60.5 & 61.2 & 61.5 & 62.5 & 62.5 \\
StackEval & 78.7 & 78.7 & 78.6 & 77.7 & 77.5 \\
MBPP+ & 81.3 & 81.1 & 79.3 & 81.1 & 80.2 \\
LiveBenchIF & 69.2 & 68.3 & 68.5 & 68.7 & 67.6 \\
\midrule
\textbf{Mean} & \textbf{71.6} & 69.8 & 70.2 & 70.0 & 70.7 \\
\bottomrule
\end{tabular}
\end{table}

\begin{table}[t]
\centering
\scriptsize
\setlength{\tabcolsep}{5.2pt}
\caption{Optimizer swap, primary subject (\%, seed 0, budget 250); \textbf{Ours} is \method.}
\label{tab:optswap}
\begin{tabular}{l ccc ccc}
\toprule
\rowcolor{headgray} & \multicolumn{3}{c}{\textbf{Opt.: GLM-5-FP8}} & \multicolumn{3}{c}{\textbf{Opt.: Qwen-3.5-27B}} \\
\cmidrule(lr){2-4} \cmidrule(lr){5-7}
\rowcolor{headgray} \textbf{Benchmark} & \textbf{Ours} & \textbf{GEPA} & \textbf{OPRO} & \textbf{Ours} & \textbf{GEPA} & \textbf{OPRO} \\
\midrule
HotpotQA  & \bfirst{68.9} & \usecond{50.2} & 49.7 & \bfirst{69.8} & \usecond{50.9} & 48.6 \\
CollIE    & \bfirst{76.5} & 68.3 & \usecond{71.5} & \usecond{70.5} & 69.5 & \bfirst{74.0} \\
IFBench   & \bfirst{64.5} & 57.5 & \usecond{64.0} & \bfirst{65.0} & 59.5 & \usecond{63.3} \\
SuperGPQA & \bfirst{63.0} & \usecond{61.5} & \usecond{61.5} & \usecond{62.5} & 61.0 & \bfirst{65.0} \\
StackEval & \bfirst{78.3} & \usecond{78.0} & 77.5 & \bfirst{78.4} & \usecond{77.6} & 75.8 \\
MBPP+     & \usecond{79.3} & \bfirst{80.0} & 77.3 & \bfirst{80.7} & \usecond{80.0} & \usecond{80.0} \\
LiveBenchIF & \bfirst{67.7} & 66.5 & \usecond{67.2} & \usecond{67.4} & \usecond{67.4} & \bfirst{67.8} \\
\midrule
\textbf{Mean} & \bfirst{71.2} & 66.0 & \usecond{67.0} & \bfirst{70.6} & 66.6 & \usecond{67.8} \\
\bottomrule
\end{tabular}
\end{table}

Pooling the 91 main runs whose final prompt beats the seed on the slice, completed rounds reach 75\% of that slice gain by half the budget on average (Figure~\ref{fig:analysis}h).
On the primary subject, with budgets up to 1\,000, \method leads at 100 calls (68.4 against 66.6) and is flat at 71.2--71.6\% from 250 on, while GEPA needs 1\,000 calls, four times as many, to match \method's 250-call score (71.9 against 71.6).

\paragraph{GEPA returns the seed when the budget is tight.}
At budget 250 on the primary subject, 71\% of GEPA runs (15 of 21) return the unchanged five-word seed, having spent the whole budget without selecting a better candidate, against 5\% for \method (1 of 21) and 33\% for OPRO (7 of 21; Figure~\ref{fig:analysis}a).
Figure~\ref{fig:analysis}(c) and Appendix~\ref{app:trajectory} trace one run candidate by candidate.
At budget 1\,000, GEPA's seed-return rate drops to 5\% and its score rises 4.8 points over budget 250, reflecting its sensitivity to available budget.
A seed return is a returned prompt identical to the seed up to surrounding whitespace, a diagnostic that needs no re-scoring: GEPA returns the seed on 86\% of its 105 runs against 13\% for \method, OPRO on 24\% and GrIPS on 16\%; GEPA's rate ranges from 71\% on Gemma-4-31B to 95\% on Ministral-8B, while \method's never exceeds 24\% on any subject (Figure~\ref{fig:teaser}b).
MIPROv2 and aPSF always rewrite, so they never return the seed; the per-subject breakdown is in Table~\ref{tab:escape}, Appendix~\ref{app:escape}.

\subsection{A3: All Four Components Contribute}
\label{sec:ablation}

In this section, we ablate components and hyper-parameters on the primary subject.  From Table~\ref{tab:design} we can observe that 1.~removing adaptive slice sizing costs 1.8 points, the reflective operator as a whole 1.6, and the fixed slice 1.4, taking the mean from 71.6\% to 69.8--70.2\%; 2.~removing concrete failure examples costs 0.9 points on the seven-benchmark mean (0.87 on the other six, 1.1 on HotpotQA); 3.~adaptive sizing and fixed slicing pay off most on noisy tasks, worth 8.2 and 8.4 points on HotpotQA, and add 3.4 and 1.5 on CollIE; 4.~the reflective operator adds 5.5 points on HotpotQA, 4.4 on CollIE, whose constraints demand explicit format, and 1.0 on StackEval: every component is worth over four points on some task.

\paragraph{Performance is stable across hyper-parameters.}
As seen in Figure~\ref{fig:sens} (per-benchmark values in Table~\ref{tab:app_sens}), test score peaks at the default $\tse=0.10$ (71.6\%, against 69.2--70.3\% for 0.05, 0.15 and 0.20), which keeps the slice rule adaptive most often (32 of 61 runs; Appendix~\ref{app:hyper}); smaller or larger targets turn it into a fixed slice, as $\tse=0.05$ pins 15 of 21 runs to the upper clamp and $\tse\geq0.15$ pins 38 of 42 to the lower one.
For candidates per step, $\ncand=2$ (71.6\%) edges out 1 (71.1\%) and 3 (70.3\%).
We also vary the probe size ($n_\text{probe}$ from 4 to 48) and lower clamp ($s_{\min}$ from 5 to 20), as shown in Figure~\ref{fig:sens}(c, d).
All main-table runs use the same default settings across the seven tasks and five subjects, without per-task tuning.


\subsection{A4: An Optimizer Swap Keeps the Gain}
\label{sec:optimizer}

In this section, we swap the default GLM-5-FP8 optimizer for Qwen-3.5-27B~\citep{yang2025qwen3} at seed 0 (Table~\ref{tab:optswap}). We can observe that 1.~\method keeps first place under both optimizers (71.2 and 70.6) and is first on more benchmarks than either baseline, so the aggregate ranking of all three methods is preserved; 2.~with the swapped optimizer, \method (70.6) still beats both baselines run with the default one (66.0, 67.0). Across the seven benchmarks, \method leads on six with GLM-5-FP8 and four with Qwen-3.5-27B; despite differences in task outcomes, its overall advantage over both GEPA and OPRO remains consistent across the two optimizer settings.

\section{Conclusion}
\label{sec:conclusion}

We introduce \method, a single-stage optimizer that adapts a fixed evaluation slice to task noise. Under tight budgets, it leads on all five subject models, significantly outperforming every baseline, with a 13\% seed-return rate versus GEPA's 86\%. On GPT-OSS-20B, GEPA needs 5$\times$ as many calls to match \method's 100-call score, while \method uses the same defaults across tasks and subject models.

\pagebreak
\section*{Limitations}
\label{sec:limitations}

Our approach has two main limitations.  First, \method depends on the capability of the optimizer LLM that reads the failing rows and writes each candidate; the quality of its rewrites bounds what the slice can select, and a less capable optimizer may yield smaller gains (\S\ref{sec:optimizer}).  Second, our experiments currently cover text benchmarks only; extending the evaluation to multimodal benchmarks is left for future work.

\section*{Ethical Considerations}

\method is a general prompt-optimization method and, like other APO methods, could be used to optimize prompts for harmful or policy-violating objectives, depending on the subject model and usage context.
Our study uses public benchmarks and one generated instruction-stacking task, and does not collect new user data or involve human subjects.
In practice, responsible use should follow applicable platform policies and safety filters, and any released code or prompts should include usage guidelines that discourage harmful applications.

\bibliography{references}

\begin{thebibliography}{43}
\providecommand{\natexlab}[1]{#1}

\bibitem[{Agrawal et~al.(2026)Agrawal, Tan, Soylu, Ziems, Khare, Opsahl-Ong, Singhvi, Shandilya, Ryan, Jiang et~al.}]{agrawal2026gepa}
Lakshya~A Agrawal, Shangyin Tan, Dilara Soylu, Noah Ziems, Rishi Khare, Krista Opsahl-Ong, Arnav Singhvi, Herumb Shandilya, Michael~J Ryan, Meng Jiang, and 1 others. 2026.
\newblock Gepa: Reflective prompt evolution can outperform reinforcement learning.
\newblock In \emph{International Conference on Learning Representations}, volume 2026, pages 8479--8565.

\bibitem[{Anand and Chattaraj(2026)}]{anand2026instructionstackingcollapsebenchmark}
Atul Anand and Sourav Chattaraj. 2026.
\newblock \href {https://arxiv.org/abs/2608.02639} {Instruction stacking collapse: A benchmark and the capability-dependent value of prompt compilation}.
\newblock \emph{Preprint}, arXiv:2608.02639.

\bibitem[{Bergstra et~al.(2011)Bergstra, Bardenet, Bengio, and K{\'e}gl}]{bergstra2011algorithms}
James Bergstra, R{\'e}mi Bardenet, Yoshua Bengio, and Bal{\'a}zs K{\'e}gl. 2011.
\newblock Algorithms for hyper-parameter optimization.
\newblock \emph{Advances in neural information processing systems}, 24.

\bibitem[{Brown et~al.(2020)Brown, Mann, Ryder, Subbiah, Kaplan, Dhariwal, Neelakantan, Shyam, Sastry, Askell et~al.}]{brown2020language}
Tom Brown, Benjamin Mann, Nick Ryder, Melanie Subbiah, Jared~D Kaplan, Prafulla Dhariwal, Arvind Neelakantan, Pranav Shyam, Girish Sastry, Amanda Askell, and 1 others. 2020.
\newblock Language models are few-shot learners.
\newblock \emph{Advances in neural information processing systems}, 33:1877--1901.

\bibitem[{Chen et~al.(2023)Chen, Dohan, and So}]{chen2023evoprompting}
Angelica Chen, David Dohan, and David So. 2023.
\newblock Evoprompting: Language models for code-level neural architecture search.
\newblock \emph{Advances in neural information processing systems}, 36:7787--7817.

\bibitem[{Cohen(2013)}]{cohen2013statistical}
Jacob Cohen. 2013.
\newblock \emph{Statistical power analysis for the behavioral sciences}.
\newblock routledge.

\bibitem[{Du et~al.(2026)Du, Yao, Ma, Wang, Zheng, Liu, Liang, Jin, Wei, Zheng et~al.}]{du2026supergpqa}
Xeron Du, Yifan Yao, Kaijing Ma, Bingli Wang, Tianyu Zheng, Minghao Liu, Yiming Liang, Xiaolong Jin, Zhenlin Wei, Chujie Zheng, and 1 others. 2026.
\newblock Supergpqa: Scaling llm evaluation across 285 graduate disciplines.
\newblock \emph{Advances in Neural Information Processing Systems}, 38.

\bibitem[{Dwork et~al.(2015)Dwork, Feldman, Hardt, Pitassi, Reingold, and Roth}]{dwork2015generalization}
Cynthia Dwork, Vitaly Feldman, Moritz Hardt, Toni Pitassi, Omer Reingold, and Aaron Roth. 2015.
\newblock Generalization in adaptive data analysis and holdout reuse.
\newblock \emph{Advances in neural information processing systems}, 28.

\bibitem[{Fernando et~al.(2023)Fernando, Banarse, Michalewski, Osindero, and Rockt{\"a}schel}]{fernando2023promptbreeder}
Chrisantha Fernando, Dylan Banarse, Henryk Michalewski, Simon Osindero, and Tim Rockt{\"a}schel. 2023.
\newblock Promptbreeder: Self-referential self-improvement via prompt evolution.
\newblock \emph{arXiv preprint arXiv:2309.16797}.

\bibitem[{Feurer et~al.(2022)Feurer, Eggensperger, Falkner, Lindauer, and Hutter}]{feurer2022auto}
Matthias Feurer, Katharina Eggensperger, Stefan Falkner, Marius Lindauer, and Frank Hutter. 2022.
\newblock Auto-sklearn 2.0: Hands-free automl via meta-learning.
\newblock \emph{Journal of Machine Learning Research}, 23(261):1--61.

\bibitem[{Guo et~al.(2024)Guo, Wang, Guo, Li, Song, Tan, Liu, Bian, and Yang}]{guo2024connecting}
Qingyan Guo, Rui Wang, Junliang Guo, Bei Li, Kaitao Song, Xu~Tan, Guoqing Liu, Jiang Bian, and Yujiu Yang. 2024.
\newblock Connecting large language models with evolutionary algorithms yields powerful prompt optimizers.
\newblock In \emph{International Conference on Learning Representations}, volume 2024, pages 34133--34156.

\bibitem[{Holm(1979)}]{holm1979simple}
Sture Holm. 1979.
\newblock A simple sequentially rejective multiple test procedure.
\newblock \emph{Scandinavian journal of statistics}, pages 65--70.

\bibitem[{Khattab et~al.(2023)Khattab, Singhvi, Maheshwari, Zhang, Santhanam, Vardhamanan, Haq, Sharma, Joshi, Moazam et~al.}]{khattab2023dspy}
Omar Khattab, Arnav Singhvi, Paridhi Maheshwari, Zhiyuan Zhang, Keshav Santhanam, Sri Vardhamanan, Saiful Haq, Ashutosh Sharma, Thomas~T Joshi, Hanna Moazam, and 1 others. 2023.
\newblock Dspy: Compiling declarative language model calls into self-improving pipelines.
\newblock \emph{arXiv preprint arXiv:2310.03714}.

\bibitem[{Kojima et~al.(2022)Kojima, Gu, Reid, Matsuo, and Iwasawa}]{kojima2022large}
Takeshi Kojima, Shixiang~Shane Gu, Machel Reid, Yutaka Matsuo, and Yusuke Iwasawa. 2022.
\newblock Large language models are zero-shot reasoners.
\newblock \emph{Advances in neural information processing systems}, 35:22199--22213.

\bibitem[{Li et~al.(2018)Li, Jamieson, DeSalvo, Rostamizadeh, and Talwalkar}]{li2018hyperband}
Lisha Li, Kevin Jamieson, Giulia DeSalvo, Afshin Rostamizadeh, and Ameet Talwalkar. 2018.
\newblock Hyperband: A novel bandit-based approach to hyperparameter optimization.
\newblock \emph{Journal of machine learning research}, 18(185):1--52.

\bibitem[{Liu et~al.(2026)Liu, Wang, Guo, Shou, and Tang}]{liu2026adaptive}
Haoyue Liu, Zhichao Wang, Yongxin Guo, Haoran Shou, and Xiaoying Tang. 2026.
\newblock Adaptive prompt structure factorization: A framework for self-discovering and optimizing compositional prompt programs.
\newblock In \emph{Proceedings of the 64th Annual Meeting of the Association for Computational Linguistics (Volume 1: Long Papers)}, pages 11690--11714.

\bibitem[{Liu et~al.(2023{\natexlab{a}})Liu, Xia, Wang, and Zhang}]{liu2023your}
Jiawei Liu, Chunqiu~Steven Xia, Yuyao Wang, and Lingming Zhang. 2023{\natexlab{a}}.
\newblock Is your code generated by chatgpt really correct? rigorous evaluation of large language models for code generation.
\newblock \emph{Advances in neural information processing systems}, 36:21558--21572.

\bibitem[{Liu(2026)}]{liu2026bettercrosscomponentinterferencellm}
Ming Liu. 2026.
\newblock \href {https://arxiv.org/abs/2605.05716} {More is not always better: Cross-component interference in llm agent scaffolding}.
\newblock \emph{Preprint}, arXiv:2605.05716.

\bibitem[{Liu et~al.(2023{\natexlab{b}})Liu, Yuan, Fu, Jiang, Hayashi, and Neubig}]{liu2023pre}
Pengfei Liu, Weizhe Yuan, Jinlan Fu, Zhengbao Jiang, Hiroaki Hayashi, and Graham Neubig. 2023{\natexlab{b}}.
\newblock Pre-train, prompt, and predict: A systematic survey of prompting methods in natural language processing.
\newblock \emph{ACM computing surveys}, 55(9):1--35.

\bibitem[{Maurer and Pontil(2009)}]{maurer2009empirical}
Andreas Maurer and Massimiliano Pontil. 2009.
\newblock Empirical bernstein bounds and sample variance penalization.
\newblock \emph{arXiv preprint arXiv:0907.3740}.

\bibitem[{Mizrahi et~al.(2024)Mizrahi, Kaplan, Malkin, Dror, Shahaf, and Stanovsky}]{mizrahi2024state}
Moran Mizrahi, Guy Kaplan, Dan Malkin, Rotem Dror, Dafna Shahaf, and Gabriel Stanovsky. 2024.
\newblock State of what art? a call for multi-prompt llm evaluation.
\newblock \emph{Transactions of the Association for Computational Linguistics}, 12:933--949.

\bibitem[{Opsahl-Ong et~al.(2024)Opsahl-Ong, Ryan, Purtell, Broman, Potts, Zaharia, and Khattab}]{opsahl2024optimizing}
Krista Opsahl-Ong, Michael~J Ryan, Josh Purtell, David Broman, Christopher Potts, Matei Zaharia, and Omar Khattab. 2024.
\newblock Optimizing instructions and demonstrations for multi-stage language model programs.
\newblock In \emph{Proceedings of the 2024 Conference on Empirical Methods in Natural Language Processing}, pages 9340--9366.

\bibitem[{Perlitz et~al.(2024)Perlitz, Bandel, Gera, Arviv, Dor, Shnarch, Slonim, Shmueli-Scheuer, and Choshen}]{perlitz2024efficient}
Yotam Perlitz, Elron Bandel, Ariel Gera, Ofir Arviv, Liat~Ein Dor, Eyal Shnarch, Noam Slonim, Michal Shmueli-Scheuer, and Leshem Choshen. 2024.
\newblock Efficient benchmarking (of language models).
\newblock In \emph{Proceedings of the 2024 Conference of the North American Chapter of the Association for Computational Linguistics: Human Language Technologies (Volume 1: Long Papers)}, pages 2519--2536.

\bibitem[{Polo et~al.(2024)Polo, Weber, Choshen, Sun, Xu, and Yurochkin}]{polo2024tinybenchmarks}
Felipe~Maia Polo, Lucas Weber, Leshem Choshen, Yuekai Sun, Gongjun Xu, and Mikhail Yurochkin. 2024.
\newblock tinybenchmarks: evaluating llms with fewer examples.
\newblock \emph{arXiv preprint arXiv:2402.14992}.

\bibitem[{Prasad et~al.(2023)Prasad, Hase, Zhou, and Bansal}]{prasad2023grips}
Archiki Prasad, Peter Hase, Xiang Zhou, and Mohit Bansal. 2023.
\newblock Grips: Gradient-free, edit-based instruction search for prompting large language models.
\newblock In \emph{Proceedings of the 17th Conference of the European Chapter of the Association for Computational Linguistics}, pages 3845--3864.

\bibitem[{Pryzant et~al.(2023)Pryzant, Iter, Li, Lee, Zhu, and Zeng}]{pryzant2023automatic}
Reid Pryzant, Dan Iter, Jerry Li, Yin Lee, Chenguang Zhu, and Michael Zeng. 2023.
\newblock Automatic prompt optimization with “gradient descent” and beam search.
\newblock In \emph{Proceedings of the 2023 conference on empirical methods in natural language processing}, pages 7957--7968.

\bibitem[{Pyatkin et~al.(2025)Pyatkin, Malik, Graf, Ivison, Huang, Dasigi, Lambert, and Hajishirzi}]{pyatkin2025generalizingverifiableinstructionfollowing}
Valentina Pyatkin, Saumya Malik, Victoria Graf, Hamish Ivison, Shengyi Huang, Pradeep Dasigi, Nathan Lambert, and Hannaneh Hajishirzi. 2025.
\newblock \href {https://arxiv.org/abs/2507.02833} {Generalizing verifiable instruction following}.
\newblock \emph{Preprint}, arXiv:2507.02833.

\bibitem[{Reynolds and McDonell(2021)}]{reynolds2021prompt}
Laria Reynolds and Kyle McDonell. 2021.
\newblock Prompt programming for large language models: Beyond the few-shot paradigm.
\newblock In \emph{Extended abstracts of the 2021 CHI conference on human factors in computing systems}, pages 1--7.

\bibitem[{Sclar et~al.(2024)Sclar, Choi, Tsvetkov, and Suhr}]{sclar2024quantifying}
Melanie Sclar, Yejin Choi, Yulia Tsvetkov, and Alane Suhr. 2024.
\newblock Quantifying language models' sensitivity to spurious features in prompt design or: How i learned to start worrying about prompt formatting.
\newblock In \emph{International Conference on Learning Representations}, volume 2024, pages 25055--25083.

\bibitem[{Shi et~al.(2024)Shi, Yang, Chen, Li, Yang, and Shen}]{shi2024efficient}
Chengshuai Shi, Kun Yang, Zihan Chen, Jundong Li, Jing Yang, and Cong Shen. 2024.
\newblock Efficient prompt optimization through the lens of best arm identification.
\newblock \emph{Advances in Neural Information Processing Systems}, 37:99646--99685.

\bibitem[{Team et~al.(2026)Team, Abd, Aggarwal, Algayres, Andreev, Bachem, Ballantyne, Brick, C{\u{a}}rbune, Casbon et~al.}]{team2026gemma}
Gemma Team, Sherif~El Abd, Vaibhav Aggarwal, Robin Algayres, Alek Andreev, Olivier Bachem, Ian Ballantyne, Cormac Brick, Victor C{\u{a}}rbune, Michelle Casbon, and 1 others. 2026.
\newblock Gemma 4 technical report.
\newblock \emph{arXiv preprint arXiv:2607.02770}.

\bibitem[{Vivek et~al.(2024)Vivek, Ethayarajh, Yang, and Kiela}]{vivek2024anchor}
Rajan Vivek, Kawin Ethayarajh, Diyi Yang, and Douwe Kiela. 2024.
\newblock Anchor points: Benchmarking models with much fewer examples.
\newblock In \emph{Proceedings of the 18th Conference of the European Chapter of the Association for Computational Linguistics (Volume 1: Long Papers)}, pages 1576--1601.

\bibitem[{Wang et~al.(2024)Wang, Li, Wang, Bai, Luo, Zhang, Jojic, Xing, and Hu}]{wang2024promptagent}
Xinyuan Wang, Chenxi Li, Zhen Wang, Fan Bai, Haotian Luo, Jiayou Zhang, Nebojsa Jojic, Eric Xing, and Zhiting Hu. 2024.
\newblock Promptagent: Strategic planning with language models enables expert-level prompt optimization.
\newblock In \emph{International Conference on Learning Representations}, volume 2024, pages 23967--24001.

\bibitem[{Wei et~al.(2022)Wei, Wang, Schuurmans, Bosma, Xia, Chi, Le, Zhou et~al.}]{wei2022chain}
Jason Wei, Xuezhi Wang, Dale Schuurmans, Maarten Bosma, Fei Xia, Ed~Chi, Quoc~V Le, Denny Zhou, and 1 others. 2022.
\newblock Chain-of-thought prompting elicits reasoning in large language models.
\newblock \emph{Advances in neural information processing systems}, 35:24824--24837.

\bibitem[{White et~al.(2025)White, Dooley, Roberts, Pal, Feuer, Jain, Shwartz-Ziv, Jain, Saifullah, Dey et~al.}]{white2025livebench}
Colin White, Samuel Dooley, Manley Roberts, Arka Pal, Benjamin Feuer, Siddhartha Jain, Ravid Shwartz-Ziv, Neel Jain, Khalid Saifullah, Sreemanti Dey, and 1 others. 2025.
\newblock Livebench: A challenging, contamination-limited llm benchmark.
\newblock In \emph{International Conference on Learning Representations}, volume 2025, pages 91595--91631.

\bibitem[{Wilcoxon(1945)}]{wilcoxon1945individual}
Frank Wilcoxon. 1945.
\newblock Individual comparisons by ranking methods.
\newblock \emph{Biometrics bulletin}, 1(6):80--83.

\bibitem[{Yang et~al.(2025)Yang, Li, Yang, Zhang, Hui, Zheng, Yu, Gao, Huang, Lv et~al.}]{yang2025qwen3}
An~Yang, Anfeng Li, Baosong Yang, Beichen Zhang, Binyuan Hui, Bo~Zheng, Bowen Yu, Chang Gao, Chengen Huang, Chenxu Lv, and 1 others. 2025.
\newblock Qwen3 technical report.
\newblock \emph{arXiv preprint arXiv:2505.09388}.

\bibitem[{Yang et~al.(2024)Yang, Wang, Lu, Liu, Le, Zhou, and Chen}]{yang2024large}
Chengrun Yang, Xuezhi Wang, Yifeng Lu, Hanxiao Liu, Quoc~V Le, Denny Zhou, and Xinyun Chen. 2024.
\newblock Large language models as optimizers.
\newblock In \emph{International Conference on Learning Representations}, volume 2024, pages 12028--12068.

\bibitem[{Yang et~al.(2018)Yang, Qi, Zhang, Bengio, Cohen, Salakhutdinov, and Manning}]{yang2018hotpotqa}
Zhilin Yang, Peng Qi, Saizheng Zhang, Yoshua Bengio, William Cohen, Ruslan Salakhutdinov, and Christopher~D Manning. 2018.
\newblock Hotpotqa: A dataset for diverse, explainable multi-hop question answering.
\newblock In \emph{Proceedings of the 2018 conference on empirical methods in natural language processing}, pages 2369--2380.

\bibitem[{Yao et~al.(2024)Yao, Chen, Hanjie, Yang, and Narasimhan}]{yao2024collie}
Shunyu Yao, Howard Chen, Austin Hanjie, Runzhe Yang, and Karthik Narasimhan. 2024.
\newblock Collie: Systematic construction of constrained text generation tasks.
\newblock In \emph{International Conference on Learning Representations}, volume 2024, pages 38595--38606.

\bibitem[{Zehle et~al.(2025)Zehle, Schlager, Hei{\ss}, and Feurer}]{zehle2025capo}
Tom Zehle, Moritz Schlager, Timo Hei{\ss}, and Matthias Feurer. 2025.
\newblock Capo: Cost-aware prompt optimization.
\newblock \emph{arXiv preprint arXiv:2504.16005}.

\bibitem[{Zeng et~al.(2026)Zeng, Lv, Hou, Du, Zheng, Chen, Yin, Ge, Huang, Xie et~al.}]{zeng2026glm}
Aohan Zeng, Xin Lv, Zhenyu Hou, Zhengxiao Du, Qinkai Zheng, Bin Chen, Da~Yin, Chendi Ge, Chenghua Huang, Chengxing Xie, and 1 others. 2026.
\newblock Glm-5: from vibe coding to agentic engineering.
\newblock \emph{arXiv preprint arXiv:2602.15763}.

\bibitem[{Zhou et~al.(2022)Zhou, Muresanu, Han, Paster, Pitis, Chan, and Ba}]{zhou2022large}
Yongchao Zhou, Andrei~Ioan Muresanu, Ziwen Han, Keiran Paster, Silviu Pitis, Harris Chan, and Jimmy Ba. 2022.
\newblock Large language models are human-level prompt engineers.
\newblock \emph{arXiv preprint arXiv:2211.01910}.

\end{thebibliography}

\clearpage
\appendix

\section{Fixed-Pair Reliability and the Search Budget}
\label{app:theory}

This appendix gives a mechanism-level analysis of the evaluation design.  The statistical result concerns a fixed candidate pair on an independent sample, and the budget identity concerns Algorithm~\ref{alg:main}; for the adaptive search as a whole, we evaluate on held-out test pools never seen during search (\S\ref{sec:main}, Appendix~\ref{app:gengap}).

\subsection{Comparison Model}
\label{app:theory_setup}

Fix an incumbent $p$, a candidate $c$, and a positive integer $s$ before observing the evaluation sample.  Let $(X_j,Y_j)_{j=1}^s$ be i.i.d. score pairs, where $X_j=f_{I_j}(c)$ and $Y_j=f_{I_j}(p)$ evaluate the two prompts on the same random row $I_j$.  Scores lie in $[0,1]$.  The probability law includes row sampling and, if the subject model is stochastic, its decoding randomness.  Within a row the two scores need not be independent; across rows the pairs are independent and identically distributed.  This idealized comparison isolates the effect of pairing; the implemented slice is discussed under \emph{Relation to the Implemented Rule} below.

Write
\begin{align}
  Z_j &= X_j-Y_j, &
  \Delta &= \mathbb{E}[Z_j], \label{eq:theory_difference}\\
  \widehat\Delta_s &= \frac{1}{s}\sum_{j=1}^s Z_j, &
  v &= \mathrm{Var}(Z_j). \label{eq:theory_variance}
\end{align}
Here $\Delta$ is the difference of population mean scores under this evaluation law, not an observed slice or test-set difference.  With $\sigma_c^2=\mathrm{Var}(X_j)$, $\sigma_p^2=\mathrm{Var}(Y_j)$ and $\gamma_{cp}=\mathrm{Cov}(X_j,Y_j)$, expansion of the variance gives
\begin{equation}
  v=\sigma_c^2+\sigma_p^2-2\gamma_{cp},\qquad
  \mathrm{Var}(\widehat\Delta_s)=\frac{v}{s}.
  \label{eq:theory_pairing}
\end{equation}
If both marginal variances are positive, $\gamma_{cp}=\rho_{cp}\sigma_c\sigma_p$, where $\rho_{cp}$ is the same-row Pearson correlation.  The covariance form also covers zero marginal variance, for which that correlation is undefined.  Independent evaluation samples with the same marginals have $\gamma_{cp}=0$.  Thus pairing lowers the variance when $\gamma_{cp}>0$, leaves it unchanged when $\gamma_{cp}=0$, and can increase it when $\gamma_{cp}<0$.

This comparison holds the number of observations per prompt fixed and covers any sign of the same-row covariance.  The Spearman correlation in Appendix~\ref{app:gengap} measures a different relationship, between search fitness and held-out performance across prompts.

\subsection{The Comparison Bound}

\begin{proposition}[Fixed-pair comparison]
\label{prop:paired_error}
If $\Delta=\mathbb{E}[\widehat\Delta_s]>0$ and scores lie in $[0,1]$, the better candidate is rejected with probability
\begin{equation}
  \Pr(\widehat\Delta_s\leq0)
  \leq \exp\!\left(-\frac{s\Delta^2}{2v+\frac43\Delta}\right).
  \label{eq:misrank}
\end{equation}
\end{proposition}

This is a Bernstein bound~\citep{maurer2009empirical}: it ties pairing to decision reliability, and inverting it prices precision in refinement steps, since $T(s)$ is non-increasing (Corollary~\ref{cor:budget_precision}).

\paragraph{Proof of Proposition~\ref{prop:paired_error}.}
Suppose $\Delta>0$ and let $W_j=\Delta-Z_j$.  Then $\mathbb{E}[W_j]=0$, $\mathrm{Var}(W_j)=v$, and $|W_j|\leq2$ because $Z_j\in[-1,1]$.  Bernstein's exponential-moment bound for bounded, centered variables yields, for $0<\lambda<3/2$,
\begin{equation}
  \log\mathbb{E}\exp(\lambda W_j)
  \leq \frac{\lambda^2v}{2(1-2\lambda/3)}.
  \label{eq:theory_mgf}
\end{equation}
For completeness, expand the exponential and use $\mathbb{E}|W_j|^m\leq2^{m-2}v$ and $m!\geq2\cdot3^{m-2}$ for $m\geq2$; summing the resulting geometric series and applying $\log(1+x)\leq x$ gives Equation~\ref{eq:theory_mgf}.  This is the standard variance-sensitive concentration argument underlying Bernstein-type bounds~\citep{maurer2009empirical}.

Independence across rows and the exponential Markov inequality now give
\begin{align}
  \Pr(\widehat\Delta_s\leq0)
  &=\Pr\!\left(\sum_{j=1}^s W_j\geq s\Delta\right) \notag\\
  &\leq\exp\!\left[-\lambda s\Delta
       +\frac{s\lambda^2v}{2(1-2\lambda/3)}\right].
  \label{eq:theory_chernoff}
\end{align}
For $v>0$, substitute $\lambda=\Delta/(v+2\Delta/3)$ to obtain Equation~\ref{eq:misrank}.  If $v=0$, each $Z_j=\Delta>0$ almost surely, so the rejection probability is zero and the bound still holds.\hfill$\square$

The same argument applied to $-Z_j$ bounds erroneous acceptance when $\Delta<0$: $\Pr(\widehat\Delta_s>0)$ is at most the right-hand side of Equation~\ref{eq:misrank} with $\Delta$ replaced by $|\Delta|$.  At $\Delta=0$ the two prompts are equally good, so either decision costs nothing in expectation.  For fixed $\Delta>0$ and $s$, positive same-row covariance tightens the bound through $v$.

\subsection{A Sufficient Precision Target and Its Budget Cost}

\begin{corollary}[Budget cost]
\label{cor:budget_precision}
Under Proposition~\ref{prop:paired_error}, fix $\delta\in(0,1)$, $\Delta>0$, and $v$.  For integer slice limits define
\begin{equation}
  s_{\mathrm{cert}}=
  \max\!\left\{s_{\min},
  \left\lceil\frac{2v+\frac43\Delta}{\Delta^2}
  \log\frac{1}{\delta}\right\rceil\right\}.
  \label{eq:slice_certificate}
\end{equation}
If $s_{\mathrm{cert}}\leq s_{\max}$, evaluating the fixed pair on $s_{\mathrm{cert}}$ i.i.d. rows makes its rejection-error bound at most $\delta$.  Under the call accounting of Equation~\ref{eq:steps}, this size also permits at least one complete refinement step if
\begin{equation}
  n_\text{probe}+(\ncand+1)s_{\mathrm{cert}}\leq B.
  \label{eq:certificate_feasibility}
\end{equation}
When both conditions hold, $s_{\mathrm{cert}}$ maximizes the number of affordable complete steps among permitted integer slice sizes satisfying this sufficient certificate.
\end{corollary}

\paragraph{Proof of Corollary~\ref{cor:budget_precision}.}
Requiring the right-hand side of Equation~\ref{eq:misrank} to be at most $\delta$ gives
\begin{equation}
  s\geq\frac{2v+\frac43\Delta}{\Delta^2}\log\frac{1}{\delta}.
  \label{eq:certificate_requirement}
\end{equation}
Equation~\ref{eq:slice_certificate} is the smallest permitted integer satisfying this sufficient inequality.  After the probe and seed scoring, one complete step costs $\ncand s$ calls, giving Equation~\ref{eq:certificate_feasibility}.  Finally,
\begin{equation}
  T(s)=\left\lfloor
      \frac{B-n_\text{probe}}{\ncand s}-\frac{1}{\ncand}
      \right\rfloor
  \label{eq:theory_steps}
\end{equation}
is non-increasing in $s$ on the feasible domain.  The smallest certified size therefore attains the largest $T(s)$ within that domain, possibly tied with larger sizes.\hfill$\square$

This is an oracle comparison: $v$ and $\Delta$ are population quantities that \method does not observe.  The certificate is sufficient rather than necessary; when $v=0$, for example, any slice size compares the pair without error.  Its role is to make the trade-off explicit: finer comparison costs refinement steps.

\subsection{Relation to the Implemented Rule}

Equation~\ref{eq:se} motivates $s\approx\sigma^2/\tse^2$ for estimating one fixed prompt's mean, while Equation~\ref{eq:slice_certificate} concerns a particular comparison and depends on its gap and difference variance.  The implemented rule therefore uses the seed-probe estimate $\hat\sigma^2$ as a practical proxy that needs no knowledge of candidates not yet written; the ablations and sweeps of \S\ref{sec:ablation} measure its effect.

There is a second distinction: Algorithm~\ref{alg:main} chooses $s$ from the probe and includes probe rows in $S$, and subsequent candidates are written from scores and failing examples on $S$.  The bound applies to each fixed candidate pair; extending it uniformly over the adaptive search would call for tools such as adaptive data analysis~\citep{dwork2015generalization}.  We instead assess the adaptive search directly on held-out test pools never seen during search (\S\ref{sec:main}, Appendix~\ref{app:gengap}).

\paragraph{Algorithmic invariants.}
Under the row-call accounting of Algorithm~\ref{alg:main}, assuming $B\geq n_\text{probe}+s$ and complete evaluation of all $\ncand$ proposals in every executed step, $t$ steps cost
\begin{equation}
  B_{\mathrm{used}}=n_\text{probe}+s+t\ncand s\leq B.
  \label{eq:theory_accounting}
\end{equation}
The loop guard establishes the inequality by induction; if termination is solely due to the guard, fewer than $\ncand s$ calls remain.  Strict acceptance also makes the incumbent's \emph{recorded slice score} non-decreasing, since a candidate replaces it only when that score increases.  Both invariants concern the search itself; its held-out effect is measured in \S\ref{sec:main}.

The analysis yields a design principle: pairing can reduce comparison variance, while finer evaluation reduces search capacity.  The ablations and budget sweeps measure how the seed-based proxy strikes this balance, with held-out diagnostics in Appendix~\ref{app:gengap}.

\onecolumn
\raggedbottom
\section{Algorithm}
\label{app:alg}

The probe and slice retain row order.  In Algorithm~\ref{alg:main}, $\Vert$ concatenates ordered lists and $[:s]$ retains the first $s$ rows.  We use $n_\text{probe}=16$ by default, $\ncand\geq1$, $\tse>0$, $|V|\geq\max\{n_\text{probe},s_{\max}\}$, and $B\geq n_\text{probe}+s_{\max}$, so calibration and seed scoring fit within the cap for any selected slice.

\begin{algorithm}[h]
\caption{\method\ (notation of \S\ref{sec:method})}
\label{alg:main}
\begin{algorithmic}[1]
\REQUIRE validation set $V$, task description $D$, seed prompt $p_0$, budget $B$, target SE $\tse$, candidates per step $\ncand$, clamp $[s_{\min}, s_{\max}]$
\STATE $P \gets n_\text{probe}$ random rows of $V$;\quad score $p_0$ on $P$
\STATE $\hat\sigma^2 \gets \frac{1}{n_\text{probe}}\sum_{i\in P}\big(f_i(p_0)-\bar f_P(p_0)\big)^2$ \hfill\COMMENT{Eq.~\ref{eq:sigma}}
\STATE $s \gets \mathrm{clamp}\big(\lfloor(\hat\sigma/\tse)^2\rceil,\, s_{\min},\, s_{\max}\big)$ \hfill\COMMENT{Eq.~\ref{eq:slice}}
\STATE $S \gets \big(P \;\Vert\; \mathrm{shuffle}(V\!\setminus\!P)\big)_{[:s]}$ \hfill\COMMENT{probe rows first}
\STATE $p \gets p_0$;\quad compute $\bar f_S(p)$;\quad $B_\text{used} \gets n_\text{probe} + s$
\WHILE{$B - B_\text{used} \geq \ncand\, s$}
  \STATE $E_S(p) \gets (\,i\in S : f_i(p)<1\,)_{[:3]}$ \hfill\COMMENT{Eq.~\ref{eq:operator}}
  \STATE $(c^{(1)},\ldots,c^{(\ncand)}) \gets G\big(D,\,p,\,\bar f_S(p),\,E_S(p)\big)$
  \FOR{$i = 1$ \textbf{to} $\ncand$}
    \STATE compute $\bar f_S(c^{(i)})$;\quad $B_\text{used} \gets B_\text{used} + s$
    \IF{$\bar f_S(c^{(i)}) > \bar f_S(p)$}
      \STATE $p \gets c^{(i)}$ \hfill\COMMENT{Eq.~\ref{eq:accept}}
    \ENDIF
  \ENDFOR
\ENDWHILE
\RETURN $p$
\end{algorithmic}
\end{algorithm}

\twocolumn
\flushbottom

\section{Experimental Setup Details}
\label{app:setup}

\paragraph{Benchmark splits.}
Table~\ref{tab:app_splits} reports the sizes of the training, validation and test pools, denoted $\mathcal{R}_\text{cal}$, $\mathcal{R}_\text{opt}$ and $\mathcal{R}_\text{test}$.  These symbols denote finite row collections, while $\mathcal{R}$ in \S\ref{sec:method} denotes the target distribution.
\method takes both the probe ($n_\text{probe}=16$) and the fixed slice from $\mathcal{R}_\text{opt}=V$; OPRO, GrIPS and MIPROv2 sample $\mathcal{R}_\text{cal}$, aPSF samples $\mathcal{R}_\text{opt}$, and GEPA uses $\mathcal{R}_\text{cal}$ as its dataset and $\mathcal{R}_\text{opt}$ as its valset.  The designated test pool is scored after optimization and outside the budget.
For IFBench, training and validation come from IF-RLVR Train, while testing uses the first 200 released IFBench examples.  HotpotQA first samples a run-seed-dependent pool and then partitions it into training, validation and test rows.  StackEval pools use seeded generation with split-specific row IDs, without content-level deduplication across pools, and are shared by all methods.

\paragraph{Task scores.}
HotpotQA uses the local token-overlap answer-F1 scorer, with answer extraction from the last \texttt{Answer:} label or the final reply line.  CollIE scores whether all requested constraints pass.  IFBench, StackEval and LiveBenchIF average the binary checks within each row.  SuperGPQA scores the extracted option letter, and MBPP+ scores whether the generated program passes its tests.  Reported percentages are 100 times the mean row score.

\begin{table}[t]
\centering
\small
\caption{Benchmark data splits (number of rows).  Identical for all five subject models.}
\label{tab:app_splits}
\begin{tabular}{l ccc}
\toprule
\textbf{Benchmark} & $|\mathcal{R}_\text{cal}|$ & $|\mathcal{R}_\text{opt}|$ & $|\mathcal{R}_\text{test}|$ \\
\midrule
HotpotQA    & 150 & 300 & 200 \\
CollIE      & 150 & 300 & 200 \\
IFBench     & 150 & 300 & 200 \\
SuperGPQA   & 150 & 300 & 200 \\
StackEval   & 150 & 300 & 200 \\
MBPP+       &  78 & 150 & 150 \\
LiveBenchIF & 100 & 150 & 150 \\
\bottomrule
\end{tabular}
\end{table}

\paragraph{Decoding parameters.}
All five subject models use temperature 0.6, top-$p$ 0.95, and a maximum output length of 2\,048 tokens.  Top-$k$ is 20 where supported; the Mistral endpoint omits this parameter.
Every method evaluated on a given subject uses the same decoding configuration.
The optimizer model (GLM-5-FP8 or Qwen-3.5-27B) is decoded with temperature 1.0 and no nucleus/top-$k$ filtering, to encourage diverse candidate generation.

\paragraph{Significance tests.}
The paired Wilcoxon signed-rank tests use the tie-corrected normal approximation, omit zero differences, and apply no continuity correction.  Differences on the $[0,1]$ scale are rounded to 12 decimal places before ranking to remove floating-point tie artifacts.

\paragraph{Model endpoints.}
All five subject models and both optimizer models are accessed through API endpoints.
GPT-OSS-20B and Nemotron-3-Super-120B are queried through the NVIDIA API, and Ministral-8B through the Mistral API.
Within each subject model, all six methods use the same configured endpoint, decoding settings and seed prompt.
All four hyper-parameter sweeps in Figure~\ref{fig:sens} use the same Gemma-4-31B API endpoint as the main experiments.

\paragraph{Budget accounting.}
The budget counts scheduled subject-model evaluations on search-time rows.
It increments by 1 for each (prompt, row) evaluation; transport retries remain part of that evaluation and are not counted as additional budget units.
The initial probe ($n_\text{probe}$ rows, 16 by default) and each candidate evaluation ($s$ rows, where $s$ is the slice size) count toward the budget.
Optimizer-model calls (prompt rewriting) are free for all methods.
Test-set evaluation at the end of each run does not count.

\paragraph{Landing guards.}
To ensure data quality, we apply two post-hoc guards to each run:
(1)~\emph{Coverage guard}: an API failure drops a row from a test evaluation silently, so a cell in which fewer than 95\% of the intended test rows were successfully scored is not written at all and is retried on the next pass.
(2)~\emph{Budget guard}: at caps of 250 calls and above, a run that spends less than 80\% of the cap (e.g., $<$200 calls out of 250) is discarded.  Below 250 no floor is applied, since a single round can legitimately leave more than 20\% unspent: at $B=100$, a 21-row slice with $\ncand=2$ stops after $16+21+42=79$ calls, too few remaining for another round.
aPSF is exempt from the budget floor.  It scores each prompt on $\min(50,\max(10,\lfloor B/10\rfloor))$ validation rows (25 at $B=250$), evaluates four candidates per update step, and stops when fewer than two such evaluations remain, after ten steps, or at a perfect validation score.  Within our budgets the first condition binds: a run makes one initial evaluation and two update steps, $9\times25=225$ calls at $B=250$ and $0.9B$ in general, unless a perfect validation score ends it earlier, as in one of the 105 main runs (125 calls).
On the primary subject model, 4 of the 126 runs were re-run due to coverage violations and 2 due to budget underutilization; the same guards were applied unchanged to the other four subject models.

\paragraph{Seed prompt.}
All methods start from the identical five-word seed prompt: ``You are a helpful assistant.''
No task-specific information is provided in the seed.

\section{Baseline Spread}
\label{app:spread}

Table~\ref{tab:app_resolution} reports, for each task, the gap between the best and worst of the five baseline means, \method excluded, next to \method's margin over the best baseline.  On the mean over tasks, the margin ($+1.98$) is larger than the spread of the five baselines (1.76).

\begin{table}[t]
\centering
\small
\caption{Baseline spread per benchmark (Table~\ref{tab:main5}b): best minus worst of the five baselines over five subjects and three seeds, \method excluded.  $\Delta$: \method's margin over the best baseline.  Last row: both quantities on the seven-task mean scores.}
\label{tab:app_resolution}
\begin{tabular}{l cc}
\toprule
\textbf{Benchmark} & \textbf{Spread (pt)} & \textbf{\method $\Delta$} \\
\midrule
MBPP+        & 6.77 & $-0.11$ \\
IFBench      & 6.73 & $-0.85$ \\
HotpotQA     & 5.19 & $+3.51$ \\
CollIE       & 5.18 & $+1.40$ \\
StackEval    & 2.68 & $+1.15$ \\
SuperGPQA    & 2.11 & $+1.15$ \\
LiveBenchIF  & 0.97 & $+0.27$ \\
\midrule
Seven-task mean & 1.76 & $+1.98$ \\
\bottomrule
\end{tabular}
\end{table}

\section{Per-Subject Per-Benchmark Results}

\label{app:per_subject}

Table~\ref{tab:main_big} gives the full per-benchmark breakdown behind the main comparison (Table~\ref{tab:main5}) for all five subject models.
Each cell is the mean over seeds 0/1/2 with the sample standard deviation across those seeds; each block's \textbf{Avg} column is the corresponding row of Table~\ref{tab:main5}.
The \emph{No opt.}\ row gives the seed prompt's score, recorded in the seed-0 \method runs.  Method comparisons use the absolute test scores of their returned prompts.
The first block, Gemma-4-31B, is the per-benchmark view of the primary subject referred to throughout \S\ref{sec:main}--\S\ref{sec:optimizer}.

Two patterns are visible across subjects.
First, on the two subjects with the widest margins \method's largest per-benchmark wins are on HotpotQA and CollIE for Gemma-4-31B ($+11.6$ and $+2.5$ over the next best) and on CollIE for GPT-OSS-20B ($+3.7$).
Second, the size of the method differences varies by task and subject: on Ministral-8B, for example, StackEval spans 57.13--70.37 points, whereas LiveBenchIF spans 49.83--54.39.  The per-cell table complements the aggregate rankings.

\begin{table*}[!t]
\centering
\scriptsize
\setlength{\tabcolsep}{2.6pt}
\caption{\textbf{Main result per cell.}  Test score (\%) at a 250-call cap, mean$_{\pm\text{std}}$ over seeds 0/1/2 (sample standard deviation); all methods share optimizer, budget, decoding and seed prompt.  \emph{No optimization}: the seed prompt at seed 0.  \textbf{Bold}: within each subject, each task column's best and every entry within one standard error of it; significance is in Table~\ref{tab:sig}.  Shaded: \method, with its margin over the best baseline in \textcolor{gaingreen}{green}.}
\label{tab:main_big}
\begin{tabular}{llccccccc|c}
\toprule
\textbf{Subject} & \textbf{Method} & \textbf{Hotpot} & \textbf{CollIE} & \textbf{IFBench} & \textbf{SuperGPQA} & \textbf{Stack} & \textbf{MBPP+} & \textbf{LB-IF} & \textbf{Avg} \\
\midrule
\midrule
\multicolumn{9}{l}{\textit{Subject = Gemma-4-31B}} \\
\addlinespace[1pt]
 & \emph{No optimization} & 50.56 & 67.50 & 57.75 & 61.00 & 77.41 & 77.33 & 67.37 & 65.56 \\
\addlinespace[1pt]
 & GEPA & 54.31{\tiny$\pm$3.91} & 69.45{\tiny$\pm$1.78} & 57.83{\tiny$\pm$1.53} & \bfirst{61.83}{\tiny$\pm$0.58} & 77.79{\tiny$\pm$0.32} & \bfirst{80.44}{\tiny$\pm$1.39} & 68.10{\tiny$\pm$1.42} & 67.11 \\
 & OPRO & 52.00{\tiny$\pm$3.00} & 71.67{\tiny$\pm$1.26} & 63.92{\tiny$\pm$1.13} & 60.50{\tiny$\pm$1.73} & 77.96{\tiny$\pm$1.02} & 79.33{\tiny$\pm$2.00} & 67.64{\tiny$\pm$0.72} & 67.58 \\
 & MIPROv2 & 49.79{\tiny$\pm$4.45} & 70.67{\tiny$\pm$0.76} & 64.33{\tiny$\pm$1.88} & \bfirst{61.50}{\tiny$\pm$0.50} & 73.38{\tiny$\pm$4.66} & 79.33{\tiny$\pm$2.67} & 67.37{\tiny$\pm$0.05} & 66.62 \\
 & GrIPS & 57.12{\tiny$\pm$1.98} & 68.11{\tiny$\pm$0.20} & 56.92{\tiny$\pm$1.84} & 61.33{\tiny$\pm$1.26} & 77.46{\tiny$\pm$0.73} & \bfirst{80.89}{\tiny$\pm$1.02} & 66.92{\tiny$\pm$0.48} & 66.97 \\
 & aPSF & 62.71{\tiny$\pm$13.33} & 71.50{\tiny$\pm$1.00} & \bfirst{65.33}{\tiny$\pm$0.14} & \bfirst{62.33}{\tiny$\pm$1.53} & 75.58{\tiny$\pm$1.48} & 76.89{\tiny$\pm$2.34} & 67.60{\tiny$\pm$0.25} & 68.85 \\
\rowcolor{ourstint} & \method & \bfirst{74.34}{\tiny$\pm$6.03} & \bfirst{74.17}{\tiny$\pm$2.02} & 62.75{\tiny$\pm$3.70} & 60.50{\tiny$\pm$2.78} & \bfirst{78.67}{\tiny$\pm$0.31} & \bfirst{81.33}{\tiny$\pm$1.76} & \bfirst{69.24}{\tiny$\pm$1.37} & \bfirst{71.57}{\tiny\textcolor{gaingreen}{+2.72}} \\
\midrule
\multicolumn{9}{l}{\textit{Subject = GPT-OSS-20B}} \\
\addlinespace[1pt]
 & \emph{No optimization} & 68.13 & 66.00 & 42.75 & 36.00 & 65.13 & 72.00 & 61.62 & 58.80 \\
\addlinespace[1pt]
 & GEPA & 70.14{\tiny$\pm$2.72} & 68.67{\tiny$\pm$1.26} & 43.92{\tiny$\pm$0.63} & 35.00{\tiny$\pm$1.32} & 65.10{\tiny$\pm$0.91} & \bfirst{76.44}{\tiny$\pm$2.78} & 61.44{\tiny$\pm$0.96} & 60.10 \\
 & OPRO & \bfirst{73.87}{\tiny$\pm$2.90} & 78.00{\tiny$\pm$3.50} & 52.83{\tiny$\pm$7.04} & \bfirst{38.33}{\tiny$\pm$2.52} & 66.42{\tiny$\pm$1.63} & 70.89{\tiny$\pm$1.68} & \bfirst{62.19}{\tiny$\pm$0.24} & 63.22 \\
 & MIPROv2 & \bfirst{76.23}{\tiny$\pm$4.53} & 78.33{\tiny$\pm$5.06} & 53.17{\tiny$\pm$5.49} & 35.83{\tiny$\pm$2.08} & \bfirst{67.72}{\tiny$\pm$0.52} & 73.56{\tiny$\pm$2.78} & 61.77{\tiny$\pm$0.84} & 63.80 \\
 & GrIPS & 71.02{\tiny$\pm$3.22} & 67.17{\tiny$\pm$4.01} & 45.67{\tiny$\pm$1.01} & 34.83{\tiny$\pm$1.61} & 64.47{\tiny$\pm$1.34} & 71.33{\tiny$\pm$1.15} & 61.54{\tiny$\pm$0.52} & 59.43 \\
 & aPSF & 72.17{\tiny$\pm$5.65} & 80.00{\tiny$\pm$1.80} & \bfirst{58.17}{\tiny$\pm$3.02} & \bfirst{38.50}{\tiny$\pm$0.50} & 64.84{\tiny$\pm$1.83} & 61.11{\tiny$\pm$14.15} & \bfirst{62.05}{\tiny$\pm$0.59} & 62.41 \\
\rowcolor{ourstint} & \method & \bfirst{74.79}{\tiny$\pm$4.25} & \bfirst{83.67}{\tiny$\pm$1.44} & 54.42{\tiny$\pm$8.18} & 38.00{\tiny$\pm$1.50} & 66.97{\tiny$\pm$1.43} & \bfirst{76.22}{\tiny$\pm$3.79} & \bfirst{62.65}{\tiny$\pm$1.05} & \bfirst{65.24}{\tiny\textcolor{gaingreen}{+1.44}} \\
\midrule
\multicolumn{9}{l}{\textit{Subject = Qwen-3.5-27B}} \\
\addlinespace[1pt]
 & \emph{No optimization} & 69.62 & 50.00 & 51.50 & 52.00 & 78.93 & 80.67 & 67.24 & 64.28 \\
\addlinespace[1pt]
 & GEPA & 75.61{\tiny$\pm$4.46} & \bfirst{53.67}{\tiny$\pm$1.04} & 53.58{\tiny$\pm$0.95} & 51.67{\tiny$\pm$1.26} & \bfirst{79.14}{\tiny$\pm$0.47} & \bfirst{81.11}{\tiny$\pm$0.38} & 68.03{\tiny$\pm$0.77} & 66.11 \\
 & OPRO & 78.02{\tiny$\pm$2.52} & 51.83{\tiny$\pm$1.04} & 55.25{\tiny$\pm$2.88} & 50.00{\tiny$\pm$1.32} & 77.84{\tiny$\pm$1.65} & 79.11{\tiny$\pm$3.29} & 68.38{\tiny$\pm$0.18} & 65.78 \\
 & MIPROv2 & \bfirst{81.65}{\tiny$\pm$4.75} & \bfirst{53.00}{\tiny$\pm$3.28} & \bfirst{58.33}{\tiny$\pm$2.70} & 51.33{\tiny$\pm$1.26} & 77.47{\tiny$\pm$0.66} & 72.44{\tiny$\pm$1.54} & \bfirst{68.90}{\tiny$\pm$0.46} & 66.16 \\
 & GrIPS & 77.31{\tiny$\pm$5.34} & 50.33{\tiny$\pm$0.76} & 55.42{\tiny$\pm$1.81} & 50.83{\tiny$\pm$2.75} & 77.72{\tiny$\pm$0.60} & 79.11{\tiny$\pm$3.91} & 68.20{\tiny$\pm$0.35} & 65.56 \\
 & aPSF & 74.23{\tiny$\pm$4.55} & 52.33{\tiny$\pm$1.04} & \bfirst{57.92}{\tiny$\pm$0.76} & 48.67{\tiny$\pm$1.76} & 75.98{\tiny$\pm$0.86} & 72.00{\tiny$\pm$0.67} & 68.14{\tiny$\pm$0.98} & 64.18 \\
\rowcolor{ourstint} & \method & 77.25{\tiny$\pm$10.00} & \bfirst{54.17}{\tiny$\pm$2.57} & 56.75{\tiny$\pm$3.03} & \bfirst{56.33}{\tiny$\pm$1.04} & 78.54{\tiny$\pm$0.56} & 80.00{\tiny$\pm$0.67} & \bfirst{68.76}{\tiny$\pm$0.88} & \bfirst{67.40}{\tiny\textcolor{gaingreen}{+1.24}} \\
\midrule
\multicolumn{9}{l}{\textit{Subject = Nemotron-3-Super-120B}} \\
\addlinespace[1pt]
 & \emph{No optimization} & 72.63 & 43.00 & 40.25 & 29.00 & 72.46 & 81.33 & 68.09 & 58.11 \\
\addlinespace[1pt]
 & GEPA & \bfirst{76.54}{\tiny$\pm$3.48} & 43.17{\tiny$\pm$1.44} & 41.92{\tiny$\pm$0.72} & 29.50{\tiny$\pm$2.00} & 71.06{\tiny$\pm$1.03} & \bfirst{79.29}{\tiny$\pm$0.67} & \bfirst{68.07}{\tiny$\pm$0.48} & 58.51 \\
 & OPRO & 76.13{\tiny$\pm$4.33} & \bfirst{44.50}{\tiny$\pm$0.50} & 48.42{\tiny$\pm$1.61} & \bfirst{37.83}{\tiny$\pm$4.04} & \bfirst{72.68}{\tiny$\pm$1.98} & 76.22{\tiny$\pm$2.34} & 67.24{\tiny$\pm$0.18} & 60.43 \\
 & MIPROv2 & \bfirst{78.54}{\tiny$\pm$3.50} & \bfirst{44.50}{\tiny$\pm$2.50} & 46.08{\tiny$\pm$5.56} & 28.17{\tiny$\pm$4.04} & 70.15{\tiny$\pm$1.23} & 77.01{\tiny$\pm$3.63} & 67.14{\tiny$\pm$0.63} & 58.80 \\
 & GrIPS & 75.11{\tiny$\pm$2.90} & 43.33{\tiny$\pm$2.47} & 45.58{\tiny$\pm$2.74} & 30.00{\tiny$\pm$2.65} & \bfirst{72.58}{\tiny$\pm$0.43} & \bfirst{78.93}{\tiny$\pm$1.09} & 66.96{\tiny$\pm$0.45} & 58.93 \\
 & aPSF & 75.07{\tiny$\pm$4.49} & \bfirst{45.17}{\tiny$\pm$4.16} & 48.00{\tiny$\pm$2.17} & \bfirst{37.07}{\tiny$\pm$2.29} & 71.51{\tiny$\pm$0.72} & 73.66{\tiny$\pm$0.93} & 66.78{\tiny$\pm$0.74} & 59.61 \\
\rowcolor{ourstint} & \method & \bfirst{78.36}{\tiny$\pm$3.66} & \bfirst{45.17}{\tiny$\pm$2.47} & \bfirst{53.00}{\tiny$\pm$1.52} & \bfirst{38.67}{\tiny$\pm$5.01} & 70.45{\tiny$\pm$1.18} & 77.40{\tiny$\pm$3.70} & 66.94{\tiny$\pm$0.24} & \bfirst{61.43}{\tiny\textcolor{gaingreen}{+1.00}} \\
\midrule
\multicolumn{9}{l}{\textit{Subject = Ministral-8B}} \\
\addlinespace[1pt]
 & \emph{No optimization} & 46.56 & 37.50 & 31.75 & 35.00 & 62.94 & 71.33 & 51.96 & 48.15 \\
\addlinespace[1pt]
 & GEPA & 46.40{\tiny$\pm$3.81} & 35.17{\tiny$\pm$2.02} & 33.33{\tiny$\pm$0.63} & 33.83{\tiny$\pm$0.76} & 63.38{\tiny$\pm$0.32} & 69.78{\tiny$\pm$1.02} & 52.94{\tiny$\pm$0.61} & 47.83 \\
 & OPRO & 57.73{\tiny$\pm$2.66} & \bfirst{40.50}{\tiny$\pm$4.92} & 32.50{\tiny$\pm$2.25} & \bfirst{34.00}{\tiny$\pm$3.12} & 64.36{\tiny$\pm$3.42} & \bfirst{71.56}{\tiny$\pm$0.77} & \bfirst{54.39}{\tiny$\pm$0.40} & 50.72 \\
 & MIPROv2 & \bfirst{62.72}{\tiny$\pm$3.88} & \bfirst{41.50}{\tiny$\pm$3.28} & 31.83{\tiny$\pm$1.77} & \bfirst{35.33}{\tiny$\pm$2.31} & 57.13{\tiny$\pm$1.84} & 70.00{\tiny$\pm$0.00} & 49.83{\tiny$\pm$6.40} & 49.76 \\
 & GrIPS & 50.20{\tiny$\pm$5.56} & 35.67{\tiny$\pm$5.77} & 32.17{\tiny$\pm$2.40} & 33.33{\tiny$\pm$3.75} & 62.35{\tiny$\pm$0.43} & 70.22{\tiny$\pm$0.77} & 52.02{\tiny$\pm$0.32} & 47.99 \\
 & aPSF & \bfirst{62.05}{\tiny$\pm$4.17} & \bfirst{41.50}{\tiny$\pm$5.77} & \bfirst{34.83}{\tiny$\pm$1.46} & \bfirst{34.33}{\tiny$\pm$0.76} & 59.85{\tiny$\pm$1.92} & 69.56{\tiny$\pm$1.02} & 52.88{\tiny$\pm$1.16} & 50.72 \\
\rowcolor{ourstint} & \method & \bfirst{61.74}{\tiny$\pm$10.66} & \bfirst{40.33}{\tiny$\pm$1.04} & 33.08{\tiny$\pm$1.53} & 33.17{\tiny$\pm$0.29} & \bfirst{70.37}{\tiny$\pm$1.99} & \bfirst{71.56}{\tiny$\pm$1.39} & 53.59{\tiny$\pm$2.09} & \bfirst{51.98}{\tiny\textcolor{gaingreen}{+1.26}} \\
\bottomrule
\end{tabular}
\end{table*}

\section{Per-Baseline Significance}
\label{app:sig}

Table~\ref{tab:sig} gives the per-baseline detail behind \S\ref{sec:main}.

\begin{table}[t]
\centering
\footnotesize
\caption{Paired Wilcoxon tests against each baseline over the 35 (subject, benchmark) cells, seeds averaged.  $\Delta$: mean margin in points.  \emph{Wins}: cells where \method scores higher.  $p$: Holm-corrected.}
\label{tab:sig}
\begin{tabular}{l cccc}
\toprule
\rowcolor{headgray} \textbf{Baseline} & $\Delta$ (pt) & Cohen $d_z$ & Wins & $p_{\text{Holm}}$ \\
\midrule
GrIPS   & +3.75 & 0.819 & 29/35 & $4.5\!\times\!10^{-5}$ \\
GEPA    & +3.59 & 0.682 & 26/35 & $3.9\!\times\!10^{-4}$ \\
OPRO    & +1.98 & 0.480 & 25/35 & $1.2\!\times\!10^{-3}$ \\
aPSF    & +2.37 & 0.577 & 24/35 & $3.1\!\times\!10^{-3}$ \\
MIPROv2 & +2.49 & 0.472 & 23/35 & $5.9\!\times\!10^{-3}$ \\
\bottomrule
\end{tabular}
\end{table}

\section{Budget Curve, Full Range}
\label{app:budget}

Table~\ref{tab:app_budget} extends Figure~\ref{fig:teaser}(a) to 450 and 500 calls: GEPA first reaches \method's 100-call score at 500 calls, and MIPROv2 ties it at 500.  Every method--budget entry contains all 21 runs.

\begin{table}[t]
\centering
\scriptsize
\setlength{\tabcolsep}{3.0pt}
\renewcommand{\arraystretch}{1.25}
\caption{\textbf{Budget curve, full range.} Mean test score (\%) over seven benchmarks and three seeds on GPT-OSS-20B. Extends Figure~\ref{fig:teaser}(a) to 500 calls. GEPA first reaches \method's 100-call score at 500 calls.}
\label{tab:app_budget}
\begin{tabular}{lccccccccc}
\toprule
\rowcolor{headgray} \textbf{Method} & 100 & 150 & 200 & 250 & 300 & 350 & 400 & 450 & 500 \\
\midrule
\method & 64.9 & 67.0 & 65.7 & 65.2 & 66.2 & 66.4 & 65.6 & 65.0 & 65.2 \\
GEPA & 59.2 & 59.3 & 59.9 & 60.1 & 60.0 & 64.3 & 64.3 & 64.4 & 65.1 \\
OPRO & 61.2 & 61.9 & 62.7 & 63.2 & 63.0 & 62.6 & 63.0 & 64.3 & 63.3 \\
MIPROv2 & 62.9 & 62.9 & 61.9 & 63.8 & 64.4 & 63.8 & 63.8 & 62.6 & 64.9 \\
GrIPS & 58.5 & 58.7 & 59.4 & 59.4 & 59.5 & 59.6 & 59.0 & 58.9 & 59.3 \\
aPSF & 65.0 & 63.5 & 65.2 & 62.4 & 65.4 & 65.0 & 63.7 & 64.1 & 65.0 \\
\bottomrule
\end{tabular}
\end{table}

The primary subject was swept over a wider range at a coarser grid, 100, 250, 300, 500 and 1\,000 calls, with \method at 68.4, 71.6, 71.5, 71.2 and 71.6 and GEPA at 66.6, 67.1, 66.9, 69.6 and 71.9.  The shape is the same as on GPT-OSS with the crossing further out: \method is flat from 250 while GEPA needs 1\,000 to match it.  This sweep was run before the budget guard was in place, and in three of its 42 \method cells at 500 and 1\,000 calls (IFBench at 500, LiveBenchIF and SuperGPQA at 1\,000) the optimizer returned no candidates and the run stopped at 266, 192 and 709 calls.  These cells are kept in the means above; excluding them moves \method to 71.3 at 500 and 71.7 at 1\,000.

\section{Seed Return by Subject}
\label{app:escape}

Table~\ref{tab:escape} breaks the seed-return rate of \S\ref{sec:budget_results} down by subject model.

\begin{table}[t]
\centering
\scriptsize
\renewcommand{\arraystretch}{1.62}
\setlength{\tabcolsep}{4.4pt}
\caption{\textbf{Seed return} (\%) by subject under a 250-call cap; 21 runs per cell, 105 per method.  MIPROv2 and aPSF always rewrite.}
\label{tab:escape}
\begin{tabular}{lcccccc}
\toprule
\rowcolor{headgray} \textbf{Method} & \textbf{Gemma} & \textbf{GPT-OSS} & \textbf{Qwen} & \textbf{Nemo.} & \textbf{Minis.} & \textbf{All} \\
\midrule
\rowcolor{ourstint} \method & 5 & 10 & 24 & 14 & 14 & 13 \\
GEPA & 71 & 81 & 90 & 90 & 95 & 86 \\
OPRO & 33 & 5 & 43 & 19 & 19 & 24 \\
GrIPS & 14 & 10 & 19 & 19 & 19 & 16 \\
MIPROv2 & 0 & 0 & 0 & 0 & 0 & 0 \\
aPSF & 0 & 0 & 0 & 0 & 0 & 0 \\
\bottomrule
\end{tabular}
\end{table}

\section{Seeds}
\label{app:seeds}

The main comparison, the ablations and the hyper-parameter sweeps use seeds 0, 1 and 2, so the main table and the ablations share the same runs.  The optimizer-swap study of Table~\ref{tab:optswap} uses seed 0 as a ranking check.

\section{Inputs}
\label{app:samestart}

Every method is given the same inputs, the seed prompt $p_0=$ ``You are a helpful assistant.'' and the task description $D$, under the same budget, optimizer, decoding settings and data splits; each then follows the search procedure of its original paper.

\section{Failure-Histogram Feedback}
\label{app:histogram}

Two variants change only what the refinement operator sees, with slice sizing, the fixed slice, greedy acceptance and budget accounting unchanged.  The first replaces one of the three failing rows with a ranked histogram of failed requirement types over the whole slice; the second keeps all three rows and adds the histogram, on IFBench, whose scorer names its requirement families.  Table~\ref{tab:app_histogram} compares each with the Full run at the same seed on the primary subject.  On HotpotQA the scorer does not name failure families, so the histogram reduces to a single total of failed checks, and the replaced row is the only concrete evidence lost.

\begin{table}[t]
\centering
\small
\caption{Failure-histogram feedback, primary subject (test score in \%, budget 250).  \emph{Replace}: the histogram replaces one of the three failing rows; \emph{Add}: it is added to all three.  Full is the main-table run at the same seeds; $\Delta$ is the variant minus Full.}
\label{tab:app_histogram}
\footnotesize
\setlength{\tabcolsep}{3.5pt}
\begin{tabular}{l l c rrr}
\toprule
 & \textbf{Benchmark} & \textbf{Seeds} & \textbf{Full} & \textbf{Variant} & $\Delta$ \\
\midrule
Replace & CollIE   & 0--2 & 74.17 & 71.83 & $-2.33$ \\
        & HotpotQA & 0--2 & 74.34 & 53.76 & $-20.58$ \\
        & IFBench  & 0--2 & 62.75 & 59.25 & $-3.50$ \\
        & \emph{nine cells} & & & & $-8.80$ \\
\midrule
Add     & IFBench  & 0--1 & 64.88 & 59.75 & $-5.13$ \\
\bottomrule
\end{tabular}
\end{table}

\section{Hyper-Parameter Sweeps per Benchmark}
\label{app:sens}

Table~\ref{tab:app_sens} lists every setting behind Figure~\ref{fig:sens}, one benchmark per row.
All four hyper-parameter sweeps use the same Gemma-4-31B API endpoint.
Each sweep varies one hyper-parameter while keeping the others at their default values, with Full as the common reference.

\begin{table*}[t]
\centering
\footnotesize
\setlength{\tabcolsep}{4pt}
\caption{Hyper-parameter sweeps per benchmark, primary subject (test score in \%, budget 250, mean over three seeds). Top: target SE $\tau$ and candidates per step $k$. Bottom: probe size $n$ ($n_{\mathrm{probe}}$) and lower clamp $s$ ($s_{\min}$, upper clamp 30). Full is the common default reference. Figure~\ref{fig:sens} plots the paired differences.}

\label{tab:app_sens}
\vspace{3pt}
\begin{tabular}{lcccccc}
\toprule
\textbf{Benchmark} & Full & $\tau{=}.05$ & $\tau{=}.15$ & $\tau{=}.20$ & $k{=}1$ & $k{=}3$ \\
\midrule
HotpotQA & 74.3 & 65.0 & 61.5 & 60.9 & 74.9 & 68.9 \\
CollIE & 74.2 & 72.8 & 70.2 & 71.2 & 71.8 & 73.0 \\
IFBench & 62.7 & 62.8 & 63.1 & 61.8 & 63.9 & 62.7 \\
SuperGPQA & 60.5 & 63.0 & 62.3 & 62.0 & 60.0 & 61.0 \\
StackEval & 78.7 & 78.1 & 78.3 & 78.5 & 77.8 & 77.7 \\
MBPP+ & 81.3 & 81.8 & 80.4 & 81.6 & 80.2 & 80.9 \\
LiveBenchIF & 69.2 & 68.4 & 68.7 & 68.8 & 68.9 & 68.1 \\
\midrule
\textbf{Mean} & 71.6 & 70.3 & 69.2 & 69.2 & 71.1 & 70.3 \\
\bottomrule
\end{tabular}

\vspace{10pt}
\begin{tabular}{lcccccccccccc}
\toprule
\textbf{Benchmark} & Full & $n{=}4$ & $n{=}8$ & $n{=}12$ & $n{=}24$ & $n{=}32$ & $n{=}48$ & $s{=}5$ & $s{=}8$ & $s{=}12$ & $s{=}15$ & $s{=}20$ \\
\midrule
HotpotQA & 74.3 & 75.8 & 60.1 & 71.2 & 67.7 & 64.2 & 76.4 & 63.4 & 56.9 & 68.3 & 58.6 & 68.3 \\
CollIE & 74.2 & 75.0 & 73.5 & 73.7 & 73.3 & 71.8 & 73.7 & 72.7 & 74.2 & 72.7 & 73.0 & 72.8 \\
IFBench & 62.7 & 60.6 & 61.2 & 63.8 & 65.8 & 63.4 & 62.1 & 59.8 & 61.4 & 65.2 & 59.5 & 62.1 \\
SuperGPQA & 60.5 & 62.7 & 61.3 & 62.3 & 60.2 & 62.5 & 62.7 & 62.5 & 62.3 & 61.5 & 60.3 & 61.0 \\
StackEval & 78.7 & 77.9 & 77.9 & 77.2 & 78.7 & 76.8 & 77.4 & 76.6 & 77.2 & 78.5 & 78.5 & 78.4 \\
MBPP+ & 81.3 & 82.2 & 81.8 & 80.4 & 81.8 & 82.4 & 82.0 & 82.7 & 81.8 & 82.9 & 81.8 & 82.7 \\
LiveBenchIF & 69.2 & 68.0 & 67.6 & 69.3 & 68.3 & 68.3 & 68.6 & 68.3 & 68.8 & 68.2 & 67.9 & 68.9 \\
\midrule
\textbf{Mean} & 71.6 & 71.7 & 69.0 & 71.1 & 70.8 & 69.9 & 71.8 & 69.4 & 68.9 & 71.0 & 68.5 & 70.6 \\
\bottomrule
\end{tabular}

\end{table*}

\section{Slice Sizes under the Target-SE Sweep}
\label{app:hyper}
\label{app:ablation}

Table~\ref{tab:app_slices} records what each $\tse$ setting did to the slice size on the primary subject, read from the run logs (21 runs per setting, 61 for the default): the default keeps Equation~\ref{eq:slice} adaptive most often (32 of 61 runs), while $\tse=0.05$ pins 15 of 21 runs to the upper clamp and $\tse\geq0.15$ pins 38 of 42 to the lower one, so those settings behave mostly as fixed-slice methods.

\begin{table}[t]
\centering
\scriptsize
\setlength{\tabcolsep}{3.5pt}
\caption{Slice sizes chosen by Equation~\ref{eq:slice} under each target SE, primary subject, 21 runs per setting (61 for the default, which includes the main-table runs).}
\label{tab:app_slices}
\begin{tabular}{l ccc c}
\toprule
$\tse$ & \textbf{Range} & \textbf{At $s_{\min}$} & \textbf{At $s_{\max}$} & \textbf{Adaptive} \\
\midrule
0.05 & 15--30 & 0/21  & 15/21 & 6/21 \\
0.10 (default) & 10--25 & 29/61 & 0/61 & 32/61 \\
0.15 & 10--11 & 17/21 & 0/21 & 4/21 \\
0.20 & 10 & 21/21 & 0/21 & 0/21 \\
\bottomrule
\end{tabular}
\end{table}

\section{Seed Variance}
\label{app:variance}

Table~\ref{tab:app_variance} reports, for each method on the primary subject, the seven-benchmark mean test score and its standard deviation across seeds 0, 1 and 2.  \method combines the highest mean with a seed-to-seed standard deviation of 0.50 points, level with GrIPS and MIPROv2 and below OPRO, GEPA and aPSF.  Per-cell standard deviations for all five subject models are in Table~\ref{tab:main_big}, and that of the five-subject mean in Table~\ref{tab:main5}(a).

\begin{table}[t]
\centering
\small
\caption{Seven-benchmark mean test score (\%) and its standard deviation across three seeds (points), budget 250, primary subject (Gemma-4-31B).}
\label{tab:app_variance}
\begin{tabular}{l cc}
\toprule
\textbf{Method} & \textbf{Mean} & \textbf{Std.} \\
\midrule
\rowcolor{ourstint} \method & 71.57 & 0.50 \\
GrIPS   & 66.97 & 0.48 \\
MIPROv2 & 66.62 & 0.50 \\
OPRO    & 67.58 & 0.91 \\
GEPA    & 67.11 & 1.38 \\
aPSF    & 68.85 & 1.86 \\
\bottomrule
\end{tabular}
\end{table}

\section{Visualizing the Refinement Trajectory}
\label{app:trajectory}

\begin{figure}[t]
  \centering
  \includegraphics[width=\columnwidth]{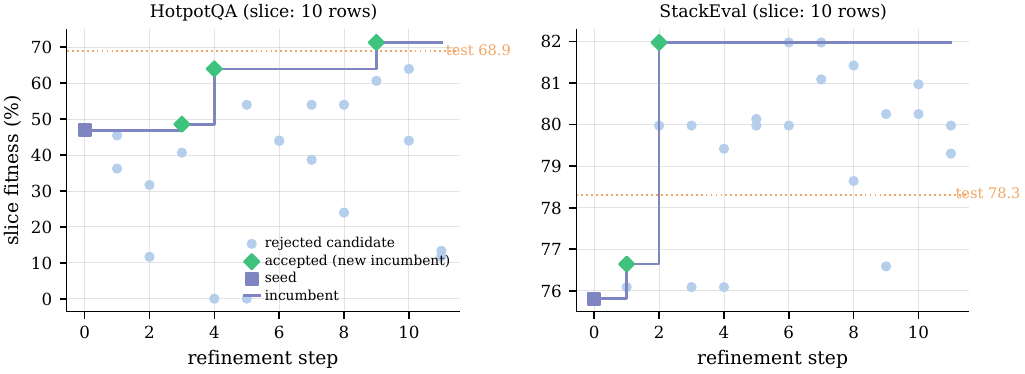}
  \caption{Every candidate of one run on HotpotQA and on StackEval (Gemma-4-31B, seed 0, budget 250): grey dots rejected, diamonds accepted, step line the incumbent.}
  \label{fig:trajectory}
\end{figure}

Figure~\ref{fig:trajectory} plots every candidate \method's search generated on one run each on HotpotQA and StackEval (Gemma-4-31B, seed 0), in the order the optimizer produced them.
The incumbent step function is exactly the greedy-accept rule of \S\ref{sec:accept}: it only moves up, and only when a candidate beats it outright.
The running best of \S\ref{sec:accept} coincides with this curve at every step: a candidate that beats the best-so-far necessarily beats the current incumbent too, because the incumbent is itself the best accepted so far.  The figure therefore shows one curve rather than two identical ones; best-tracking is a safety net for search variants that can revert the incumbent; the accept rule here never triggers it.

Both runs use the minimum slice, $s=10$, and therefore 11 steps.
On HotpotQA the proposals swing between 0\% and 71\% with no visible trend across adjacent steps, and the incumbent accepts three times, reaching 48.6\%, 64.0\% and finally 71.4\% by step 9 of 11.
On StackEval the proposals cluster inside an eight-point band from the first step onward; the incumbent locks in at step 2 of 11, at 82.0\%, and the remaining steps confirm it.

\newcommand{\gengapCells}{35}
\newcommand{\gengapDefined}{34}
\newcommand{\gengapMean}{+0.14}
\newcommand{\gengapMedian}{+0.20}
\newcommand{\gengapPositive}{21}
\newcommand{\gengapStrongPositive}{13}
\newcommand{\gengapStrongNegative}{4}
\newcommand{\gengapMinCandidates}{9}
\newcommand{\gengapMaxCandidates}{13}

\section{Generalization Gap: Slice Fitness vs.\ Held-Out Score}
\label{app:gengap}

\begin{figure*}[t]
  \centering
  \includegraphics[width=\textwidth]{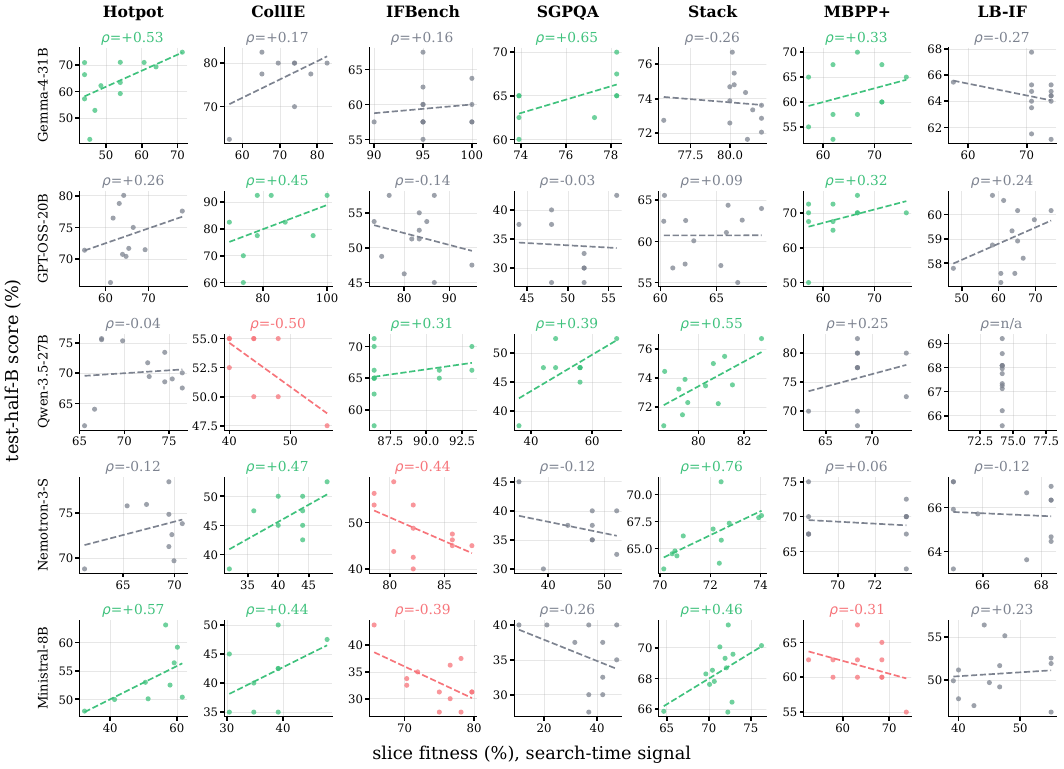}
  \caption{Slice fitness against test-half-B score over \gengapCells\ (subject, benchmark) cells, one seed-0 \method run per panel.  Spearman $\rho$: green $\geq0.3$, red $\leq-0.3$, gray otherwise, n/a for a constant slice score; dashed lines are least-squares fits.}
  \label{fig:gengap}
\end{figure*}

The slice fitness that drives every accept/reject decision in \S\ref{sec:accept} is itself an estimate, computed on $s$ rows sampled once from validation.
Figure~\ref{fig:gengap} compares that estimate with scores on the designated test pool.
For each covered (subject, benchmark) cell, we recover the seed-0 run's candidate history, retain up to 12 distinct prompts with the highest recorded slice fitness, and include the seed and shipped prompt if absent. This yields $n=\gengapMinCandidates$--$\gengapMaxCandidates$ scored prompts per cell across \gengapCells\ cells. Every candidate is evaluated on the same first 80 test rows, partitioned into alternating 40-row halves A and B. Figure~\ref{fig:gengap} uses half B for both the plotted scores and Spearman $\rho$. These rows are drawn from the main-table test pools; Appendix~\ref{app:setup} specifies their construction.

The selected prompt's test-half-B score exceeds the candidate-pool mean by a median of 1.7 points across the 35 cells, with 24 positive differences.  Relative to the seed re-scored on the same rows, its mean gain is 3.9 points and its median gain is 2.4 points.  Across the \gengapDefined\ cells with a defined correlation the median is $\rho=\gengapMedian$, with \gengapPositive\ cells positive and \gengapStrongPositive\ at $\rho\geq0.3$: the prompt chosen on the slice keeps its edge on rows never used in search.

\section{How Much Test Data Does an APO Comparison Need?}
\label{app:xl}

The main-table test pools contain 150--200 rows per run.  For each task, seed and baseline in the re-scoring study, we compute the sample standard deviation of the row-wise score differences from \method and divide by the square root of the number of jointly scored rows.  The median of these 60 paired standard errors (four tasks, three seeds, five baselines) is 2.13 points on the original rows.  This row-level quantity differs from variation across optimization seeds: Table~\ref{tab:main_big} has a median sample standard deviation of 1.61 points across three seeds, corresponding to a median standard error of 0.93 points.

We re-scored the primary subject's shipped prompts on 800 additional rows each for HotpotQA, SuperGPQA and StackEval, and 531 for CollIE.  The enlarged scores combine the original and additional evaluations, weighted by their row counts, giving totals of 1\,000 and 731 rows respectively.  Prompts are read from disk without re-optimization, and test evaluation remains outside the search budget.
Additional rows exclude the row keys used in the training, validation and test pools at seeds 0/1/2, and are shared across methods and seeds.  The row key uses up to the first 400 characters of the request, or the constraint list for StackEval.  IFBench, MBPP+ and LiveBenchIF retain their original test evaluations in this study.

\begin{table}[t]
\centering
\small
\caption{Shipped prompts re-scored on an enlarged test set (primary subject, three seeds, no re-optimization); scores pool the original and the added rows, which are shared by all methods.  $\Delta$: \method's margin over the strongest baseline.}
\label{tab:app_xl}
\setlength{\tabcolsep}{4.5pt}
\begin{tabular}{l rr rr}
\toprule
\textbf{Benchmark} & \textbf{Rows} & \textbf{Total} & $\Delta$ & $\Delta$ \textbf{enlarged} \\
\midrule
HotpotQA & 200 & 1000 & $+11.63$ & $+9.92$ \\
CollIE & 200 & 731 & $+2.50$ & $+1.46$ \\
SuperGPQA & 200 & 1000 & $-1.83$ & $-1.44$ \\
StackEval & 200 & 1000 & $+0.70$ & $+0.64$ \\
\bottomrule
\end{tabular}
\end{table}

Table~\ref{tab:app_xl} reports the original and enlarged-test comparisons.  The median paired standard error falls from 2.13 to 1.01 points, a $2.11\times$ reduction.  \method retains its position on all four tasks and stays first on HotpotQA, CollIE and StackEval; on HotpotQA its margin over the strongest baseline moves from $+11.63$ to $+9.92$.

\onecolumn
\raggedbottom
\section{Case Study: Generated Prompts}
\label{app:casestudy}

We show the exact final prompts returned at seed 0 and budget 250 with GLM-5-FP8 as optimizer on the primary subject.  The four tasks cover multi-hop QA, code, instruction following and multiple-choice knowledge; their three-seed method comparisons are in Table~\ref{tab:main_big}.

\begin{promptbox}{Seed prompt (all benchmarks)}
\small\ttfamily
You are a helpful assistant.
\end{promptbox}

\begin{promptbox}{HotpotQA (optimized prompt, 220 words, test 68.9\%)}
\raggedright\small\ttfamily
You answer reading comprehension questions based on ten provided paragraphs (two relevant, eight distractors). Your reply is scored automatically by token overlap with a short gold answer span. Any extra words in your reply cost points.

\promptheading{STRATEGY:} \\
1. Scan the ten paragraphs for the exact entity, date, name, or phrase that directly answers the question. Identify the two relevant paragraphs and ignore the eight distractors. \\
2. Extract the shortest possible verbatim substring from the text that answers the question. \\
3. Do not paraphrase, fix grammar, or expand abbreviations. If the answer is a date, output only the exact date span as it appears. If no answer is present, output the single word "No" (if "No" is incorrect, output your best guess as a short verbatim span).

\promptheading{FORMAT:} \\
Output exactly one verbatim answer span and absolutely nothing else. \\
- Do not output reasoning, preamble, or any template words. \\
- Do not include any prefix such as "Answer:". \\
- Do not repeat the answer (e.g., do not output "15 December Answer: 15 December"). Output the span exactly once. \\
- Do not output HTML or markup tags (e.g., do not use <span>). \\
- Do not output trailing punctuation, periods, or commas unless they are part of the extracted span. \\
- Do not output multiple options or variations; choose the single best verbatim span.
\end{promptbox}

This prompt asks the subject to ground its answer in the provided passages and output a short answer span.  Its formatting instructions address answer extraction and the precision term of token-overlap F1.

\clearpage
\begin{promptbox}{MBPP+ (optimized prompt, 204 words, test 79.3\%)}
\raggedright\small\ttfamily
You are a coding assistant. Your task is to output a Python function that passes hidden unit tests.

\promptheading{\#\# Strategy} \\
1. \promptlabel{Infer the True Intent}: The prompt's description is the primary specification. The example asserts only show a fraction of the required behavior. Implement the logic that fully solves the described problem, rather than just satisfying the visible asserts. \\
2. \promptlabel{Match Signatures}: Define the function with the exact name and parameter order shown in the asserts. \\
3. \promptlabel{Match Output Types}: Ensure the return type matches the example exactly (e.g., tuple vs list, string vs int). \\
4. \promptlabel{Handle Edge Cases}: Write general solutions that work for any valid input. Explicitly handle: \\
   - Empty inputs (e.g., `[]`, `""`, `()`) \\
   - Single-element inputs \\
   - Negative numbers, zeros, and boundary conditions \\
   - Unsorted or duplicate data \\
5. \promptlabel{No External Dependencies}: The code must be self-contained. Do not rely on imports or variables not defined in your reply. \\
6. \promptlabel{Avoid Hardcoding}: Do not hardcode expected outputs from the examples. Write algorithms that generalize. \\
7. \promptlabel{Verify Against Examples}: Before finalizing, mentally run the provided asserts against your code. If they fail, fix the code.

\promptheading{\#\# Format} \\
Output ONLY a valid Python code block. \\
- The code block must start with
\end{promptbox}

The saved prompt ends at the displayed line and scores 79.3\% at seed 0.  Its instructions specify function signatures, return types and edge cases.  Across three seeds, \method scores 81.33\% on MBPP+, compared with 80.89\% for the strongest baseline (Table~\ref{tab:main_big}).

\clearpage
\begin{promptbox}{IFBench (optimized prompt, 249 words, test 64.5\%)}
\raggedright\small\ttfamily
You are a helpful assistant. Your responses are evaluated by an automated system that checks if specific output constraints are met. To achieve the highest score, you must prioritize satisfying every constraint exactly as requested.

\promptheading{Strategy:} \\
1. Scan the user's entire request for explicit output constraints (e.g., word counts, required/forbidden words, case, formatting, highlighting, sentence structure). \\
2. Before generating the final response, solve the underlying task (such as the math problem or text translation). \\
3. Apply the requested constraints to the solution text only during the final drafting phase to ensure they are perfectly implemented without interfering with your reasoning.

\promptheading{Format:} \\
1. Do not include any introductory or concluding remarks unless explicitly requested. Output only the constrained text. \\
2. Do not add unprompted formatting. Unless the user explicitly requests code blocks, JSON, or markdown, output plain text. If a casing constraint is specified, apply it to the entire output, including any JSON keys, values, or structural text. \\
3. Ensure every constraint is applied globally and exactly. For example: \\
   - If asked to use all capital letters, the ENTIRE response must be in capitals. \\
   - If asked to highlight numbers, every number must be wrapped in the requested syntax (e.g., *number* or [number]). \\
   - If forbidden words are specified, ensure they do not appear anywhere in your output. \\
4. Verify constraints against your final text before sending. For format constraints, perform a direct verification of each requirement (e.g., count sentences, check case, look for forbidden terms) immediately prior to outputting.
\end{promptbox}

Step~2 (``solve the underlying task \dots\ before'' applying format constraints) has the model solve the task first and apply the format constraints afterwards, matching IFBench's dual demand for a correct answer under constraints that can conflict with producing one.

\clearpage
\begin{promptbox}{SuperGPQA (optimized prompt, 126 words, test 63.0\%)}
\raggedright\small\ttfamily
You are an expert assistant answering multiple-choice questions with options A-J. Exactly one option is correct.

Your task: analyze the problem, select the correct option, and output it in a strict format.

\promptheading{Strategy:} \\
- Identify all given numerical data, formulas, and conditions in the question. \\
- Perform the required derivation or calculation step by step using the provided values. \\
- Compare your derived result to every option A-J and select the single best match. \\
- If the question is conceptual, evaluate each option against the stated definitions and conditions.

\promptheading{Format:} \\
- Begin with a concise solution showing key steps. \\
- End with exactly one line in this format: \\
  Answer: X \\
- X must be a single letter A-J. Do not add any text after that line.
\end{promptbox}

The prompt pairs a derivation strategy (extract the given data, derive step by step, compare the result with every option, and check conceptual options against their definitions) with a strict output format that ends in a single \texttt{Answer: X} line.

\clearpage
\section{Generated Prompts: Remaining Benchmarks}
\label{app:casestudy_rest}

Appendix~\ref{app:casestudy} shows four of the seven benchmarks.  For completeness the other three are below, same run: primary subject, seed 0, budget 250, GLM-5-FP8 as optimizer.  Every prompt is reproduced exactly as the run shipped it.

\begin{promptbox}{CollIE (optimized prompt, 417 words, test 76.5\%)}
\small\ttfamily\raggedright\sloppy
Your sole objective is to satisfy exact structural constraints in user requests. A programmatic checker will parse your output; any deviation scores 0.

\medskip
\promptheading{Strategy:}\\
1. Identify every constraint: exact word counts, sentence counts, character counts, word/character positions, required/forbidden words, and specific first/last sentences or words.\\
2. Derive the exact counts mathematically before generating text. Do not approximate.\\
3. Track sentence boundaries carefully: a sentence ends with a period, question mark, or exclamation point.\\
4. Verify positions exactly: If the 4th, 8th, and 11th words must be specific words, write the text and count the words strictly. Do not add or omit punctuation that changes counts. Treat words separated by spaces as individual words.\\
5. If a specific sentence or phrase is required, copy it character-for-character, preserving exact spacing and non-ASCII characters (e.g., \textbackslash{}xa0).\\
6. Character constraints apply to the length of the entire text or a single word. Count every letter, number, and symbol. If asked for a word with exactly 12 characters where the 4th, 8th, and 11th characters are 'l', 'a', 'a', construct a 12-character string and index it. For example: "\_\_\_l\_\_\_a\_\_a\_" (replace underscores with any characters to reach exactly 12). Verify the total length and the 1-based indices before outputting.

\medskip
\promptheading{Word Counting \& Positioning:}\\
1. The checker splits text by spaces to count words. A comma or period attached to a word (e.g., "word,") counts as part of that word.\\
2. Punctuation attached to required words causes failures. If the 4th word must be "appealing", do not write "appealing,". The 4th space-separated token must be exactly "appealing".\\
3. If a position constraint seems to break natural grammar, prioritize the exact constraint. For example, if the 4th word must be "appealing", write: "The new design appealing is..." Do not force incorrect words into the required slots.\\
4. Mentally substitute required words into their exact positions first, then fill the remaining slots with placeholder words to meet the exact count. Example for 15 words (4th='for', 8th='sports', 11th='boxing'): "One two three for five six seven sports nine ten boxing twelve thirteen fourteen fifteen." Adjust the placeholders to form a sentence, but never alter the required words or their positions. Verify the final space-separated token indices before finishing.

\medskip
\promptheading{Format:}\\
1. Output strictly the requested text.\\
2. Do not include any preamble, conversational filler, labels, explanations, or quotation marks unless they are explicitly requested as part of the text.\\
3. Begin your response immediately with the first word of the requested text and end immediately after the final character.
\end{promptbox}

\clearpage
\begin{promptbox}{StackEval (optimized prompt, 162 words, test 78.3\%)}
\small\ttfamily\raggedright\sloppy
You are an automated task-solving assistant. Your output will be scored by a program that checks for 1 to 20 strict constraints related to formatting, length, wording, structure, required sections, and content.

\medskip
To maximize your score, follow these exact steps:

\medskip
1. \promptlabel{Analyze Constraints}: Scan the user request for a list of constraints. Treat every formatting and length requirement literally (e.g., exact starting words, max word count, specific keys in a JSON object, Markdown formatting).\\
2. \promptlabel{Plan Content}: Derive the core answer by directly addressing the prompt's topic, ensuring the content is relevant and complete without adding unnecessary filler.\\
3. \promptlabel{Execute Formatting}: Construct your response to satisfy all structural constraints simultaneously. Do not add conversational filler, introductory text, or labels like "answer:" unless explicitly requested.\\
4. \promptlabel{Validate}: Before finalizing, re-read the requested constraints and verify your output matches them exactly.

\medskip
\promptheading{\promptlabel{Format Execution Rules}:}\\
- If a JSON output is required, output only valid, parseable JSON. Do not include Markdown code blocks (e.g., no
\end{promptbox}

\clearpage
\begin{promptbox}{LiveBenchIF (optimized prompt, 272 words, test 67.7\%)}
\small\ttfamily\raggedright\sloppy
You are an instruction-following assistant evaluated by code. Each prompt contains a short writing task and verifiable constraints (e.g., required/forbidden keywords, start/end phrases, word/character counts, casing). Your score is the fraction of constraints satisfied. Optimize strictly for this score.

\medskip
Follow this process:

\medskip
1. ANALYSIS: Read the entire prompt. Identify:\\
   - The writing task (paraphrase, summary, story, etc.) and any source text.\\
   - Every verifiable constraint. Enumerate them internally:\\
     * Required words/phrases (must be verbatim, exact case).\\
     * Forbidden words/phrases (never appear in any form).\\
     * Exact start/end phrases (must be the very first/last characters).\\
     * Length limits (word count, character count, sentence count).\\
     * Casing rules.\\
   - Trap doors: the prompt may include a short, bland request followed by detailed constraints or source text that materially changes the task; always parse the entire prompt before drafting.

\medskip
2. DRAFTING: Write only the requested content. Do not include preambles, explanations, conversational filler, or notes. If the task provides source text, ground your answer in it. When constraints conflict with fluency, obey constraints.

\medskip
3. FINAL QA CHECKS (perform all before finalizing):\\
   - Required words: Search your draft for each exact required word/phrase. If missing, add it.\\
   - Forbidden words: Search your draft for each forbidden word in any casing. If present, rephrase to eliminate it.\\
   - Start/end: If an exact phrase is required, verify it is literally the first/last characters (including punctuation and quotes).\\
   - Length: Count words/characters as specified and trim or expand within the limit.\\
   - Casing: Apply the requested casing (e.g., all lowercase, title case).

\medskip
4. OUTPUT: Emit only the final response text, with no extra whitespace, trailing newlines, or wrapper formatting.
\end{promptbox}

\clearpage
\section{A Run, Prompt by Prompt}
\label{app:trajectory_text}

This appendix prints every accepted candidate from four \method{} runs (primary subject, seed 0, budget 250), with its recorded slice fitness.  Figure~\ref{fig:trajectory} plots the HotpotQA and StackEval runs.  Across the four runs, 11 of the 62 evaluated candidates were accepted and 51 were rejected.  The headings number accepted candidates; candidate IDs retain the search-round number.

\subsection{HotpotQA}

Slice fitness climbs from 46.9\% to 71.4\% over 3 accepted rewrites, and the prompt that ends the run scores 68.9\% on the held-out split against 50.6\% for the seed.

\begin{promptbox}{Step 0, seed prompt (slice fitness 46.9\%)}
\small\ttfamily\raggedright\sloppy
You are a helpful assistant.
\end{promptbox}

\begin{promptbox}{Step 1, accepted candidate c3\_0 (slice fitness 48.6\%)}
\small\ttfamily\raggedright\sloppy
You answer reading comprehension questions based on ten provided paragraphs (two relevant, eight distractors). Your reply is scored automatically by token overlap with a short gold answer span. Any extra words in your reply cost points.

\medskip
\promptheading{STRATEGY:}\\
1. Scan all paragraphs for the exact entity, date, name, or phrase that directly answers the question. Ignore irrelevant context.\\
2. Extract the shortest possible verbatim substring from the text. Do not paraphrase, fix grammar, or expand abbreviations.

\medskip
\promptheading{FORMAT:}\\
Output only the verbatim answer span. No reasoning, no preamble, no HTML/markup tags, no "Answer:" prefix, no trailing punctuation unless it is part of the extracted span.
\end{promptbox}

\clearpage
\begin{promptbox}{Step 2, accepted candidate c4\_0 (slice fitness 64.0\%)}
\small\ttfamily\raggedright\sloppy
You answer reading comprehension questions based on ten provided paragraphs (two relevant, eight distractors). Your reply is scored automatically by token overlap with a short gold answer span. Any extra words in your reply cost points.

\medskip
\promptheading{STRATEGY:}\\
1. Scan all paragraphs for the exact entity, date, name, or phrase that directly answers the question. Ignore irrelevant context.\\
2. Extract the shortest possible verbatim substring from the text. Do not paraphrase, fix grammar, or expand abbreviations.

\medskip
\promptheading{FORMAT:}\\
Output exactly one verbatim answer span and absolutely nothing else.\\
- No reasoning, no preamble.\\
- No "Answer:" prefix. Do not repeat the answer (e.g., do not output "15 December Answer: 15 December").\\
- No HTML or markup tags (e.g., do not use <span>).\\
- No trailing punctuation unless it is part of the extracted span.
\end{promptbox}

\clearpage
\begin{promptbox}{Step 3, accepted candidate c9\_0 (slice fitness 71.4\%)}
\small\ttfamily\raggedright\sloppy
You answer reading comprehension questions based on ten provided paragraphs (two relevant, eight distractors). Your reply is scored automatically by token overlap with a short gold answer span. Any extra words in your reply cost points.

\medskip
\promptheading{STRATEGY:}\\
1. Scan the ten paragraphs for the exact entity, date, name, or phrase that directly answers the question. Identify the two relevant paragraphs and ignore the eight distractors.\\
2. Extract the shortest possible verbatim substring from the text that answers the question.\\
3. Do not paraphrase, fix grammar, or expand abbreviations. If the answer is a date, output only the exact date span as it appears. If no answer is present, output the single word "No" (if "No" is incorrect, output your best guess as a short verbatim span).

\medskip
\promptheading{FORMAT:}\\
Output exactly one verbatim answer span and absolutely nothing else.\\
- Do not output reasoning, preamble, or any template words.\\
- Do not include any prefix such as "Answer:".\\
- Do not repeat the answer (e.g., do not output "15 December Answer: 15 December"). Output the span exactly once.\\
- Do not output HTML or markup tags (e.g., do not use <span>).\\
- Do not output trailing punctuation, periods, or commas unless they are part of the extracted span.\\
- Do not output multiple options or variations; choose the single best verbatim span.
\end{promptbox}

\clearpage
\subsection{CollIE}

Slice fitness climbs from 56.5\% to 82.6\% over 4 accepted rewrites, and the prompt that ends the run scores 76.5\% on the held-out split against 67.5\% for the seed.

\begin{promptbox}{Step 0, seed prompt (slice fitness 56.5\%)}
\small\ttfamily\raggedright\sloppy
You are a helpful assistant.
\end{promptbox}

\begin{promptbox}{Step 1, accepted candidate c1\_0 (slice fitness 69.6\%)}
\small\ttfamily\raggedright\sloppy
Your sole objective is to satisfy exact structural constraints in user requests. A programmatic checker will parse your output; any deviation scores 0.

\medskip
\promptheading{Strategy:}\\
1. Identify every constraint: exact word counts, sentence counts, word positions, required/forbidden words, and specific first/last sentences or words.\\
2. Derive the exact counts mathematically before generating text. Do not approximate.\\
3. Track sentence boundaries carefully: a sentence ends with a period, question mark, or exclamation point.\\
4. Verify positions exactly: If the 4th, 8th, and 11th words must be specific words, write the text and count the words strictly. Do not add or omit punctuation that changes counts. Treat words separated by spaces as individual words.\\
5. If a specific sentence or phrase is required, copy it character-for-character, preserving exact spacing and non-ASCII characters (e.g., \textbackslash{}xa0).

\medskip
\promptheading{Format:}\\
1. Output strictly the requested text.\\
2. Do not include any preamble, conversational filler, labels, explanations, or quotation marks unless they are explicitly requested as part of the text.\\
3. Begin your response immediately with the first word of the requested text and end immediately after the final character.
\end{promptbox}

\clearpage
\begin{promptbox}{Step 2, accepted candidate c2\_0 (slice fitness 73.9\%)}
\small\ttfamily\raggedright\sloppy
Your sole objective is to satisfy exact structural constraints in user requests. A programmatic checker will parse your output; any deviation scores 0.

\medskip
\promptheading{Strategy:}\\
1. Identify every constraint: exact word counts, sentence counts, word positions, required/forbidden words, and specific first/last sentences or words.\\
2. Derive the exact counts mathematically before generating text. Do not approximate.\\
3. Track sentence boundaries carefully: a sentence ends with a period, question mark, or exclamation point.\\
4. Verify positions exactly: If the 4th, 8th, and 11th words must be specific words, write the text and count the words strictly. Do not add or omit punctuation that changes counts. Treat words separated by spaces as individual words.\\
5. If a specific sentence or phrase is required, copy it character-for-character, preserving exact spacing and non-ASCII characters (e.g., \textbackslash{}xa0).\\
6. Write grammatically correct sentences. Do not force required words into positions that break normal grammar (e.g., avoiding missing verbs or misplaced adjectives).\\
7. Count words as any continuous sequence of non-whitespace characters. Punctuation attached to a word (e.g., "book,") counts as part of that word and does not change its position.\\
8. After drafting your text, manually index every word from 1 to N to verify exact positions and total count. For example, if the 4th, 8th, and 11th words must be 'for', 'sports', 'boxing', verify: 1: The 2: goal 3: is 4: for 5: ... 8: sports 9: ... 11: boxing ... N. Adjust phrasing until the grammar is natural and the word counts align perfectly.

\medskip
\promptheading{Format:}\\
1. Output strictly the requested text.\\
2. Do not include any preamble, conversational filler, labels, explanations, or quotation marks unless they are explicitly requested as part of the text.\\
3. Begin your response immediately with the first word of the requested text and end immediately after the final character.
\end{promptbox}

\clearpage
\begin{promptbox}{Step 3, accepted candidate c2\_1 (slice fitness 78.3\%)}
\small\ttfamily\raggedright\sloppy
Your sole objective is to satisfy exact structural constraints in user requests. A programmatic checker will parse your output; any deviation scores 0.

\medskip
\promptheading{Strategy:}\\
1. Identify every constraint: exact word counts, sentence counts, word positions, required/forbidden words, and specific first/last sentences or words.\\
2. Derive the exact counts mathematically before generating text. Do not approximate.\\
3. Track sentence boundaries carefully: a sentence ends with a period, question mark, or exclamation point.\\
4. Verify positions exactly: If the 4th, 8th, and 11th words must be specific words, write the text and count the words strictly. Do not add or omit punctuation that changes counts. Treat words separated by spaces as individual words.\\
5. If a specific sentence or phrase is required, copy it character-for-character, preserving exact spacing and non-ASCII characters (e.g., \textbackslash{}xa0).

\medskip
\promptheading{Word Counting \& Positioning:}\\
1. The checker splits text by spaces to count words. A comma or period attached to a word (e.g., "word,") counts as part of that word.\\
2. Punctuation attached to required words causes failures. If the 4th word must be "appealing", do not write "appealing,". The 4th space-separated token must be exactly "appealing".\\
3. If a position constraint seems to break natural grammar, prioritize the exact constraint. For example, if the 4th word must be "appealing", write: "The new design appealing is..." Do not force incorrect words into the required slots.

\medskip
\promptheading{Format:}\\
1. Output strictly the requested text.\\
2. Do not include any preamble, conversational filler, labels, explanations, or quotation marks unless they are explicitly requested as part of the text.\\
3. Begin your response immediately with the first word of the requested text and end immediately after the final character.
\end{promptbox}

\clearpage
\begin{promptbox}{Step 4, accepted candidate c3\_0 (slice fitness 82.6\%)}
\small\ttfamily\raggedright\sloppy
Your sole objective is to satisfy exact structural constraints in user requests. A programmatic checker will parse your output; any deviation scores 0.

\medskip
\promptheading{Strategy:}\\
1. Identify every constraint: exact word counts, sentence counts, character counts, word/character positions, required/forbidden words, and specific first/last sentences or words.\\
2. Derive the exact counts mathematically before generating text. Do not approximate.\\
3. Track sentence boundaries carefully: a sentence ends with a period, question mark, or exclamation point.\\
4. Verify positions exactly: If the 4th, 8th, and 11th words must be specific words, write the text and count the words strictly. Do not add or omit punctuation that changes counts. Treat words separated by spaces as individual words.\\
5. If a specific sentence or phrase is required, copy it character-for-character, preserving exact spacing and non-ASCII characters (e.g., \textbackslash{}xa0).\\
6. Character constraints apply to the length of the entire text or a single word. Count every letter, number, and symbol. If asked for a word with exactly 12 characters where the 4th, 8th, and 11th characters are 'l', 'a', 'a', construct a 12-character string and index it. For example: "\_\_\_l\_\_\_a\_\_a\_" (replace underscores with any characters to reach exactly 12). Verify the total length and the 1-based indices before outputting.

\medskip
\promptheading{Word Counting \& Positioning:}\\
1. The checker splits text by spaces to count words. A comma or period attached to a word (e.g., "word,") counts as part of that word.\\
2. Punctuation attached to required words causes failures. If the 4th word must be "appealing", do not write "appealing,". The 4th space-separated token must be exactly "appealing".\\
3. If a position constraint seems to break natural grammar, prioritize the exact constraint. For example, if the 4th word must be "appealing", write: "The new design appealing is..." Do not force incorrect words into the required slots.\\
4. Mentally substitute required words into their exact positions first, then fill the remaining slots with placeholder words to meet the exact count. Example for 15 words (4th='for', 8th='sports', 11th='boxing'): "One two three for five six seven sports nine ten boxing twelve thirteen fourteen fifteen." Adjust the placeholders to form a sentence, but never alter the required words or their positions. Verify the final space-separated token indices before finishing.

\medskip
\promptheading{Format:}\\
1. Output strictly the requested text.\\
2. Do not include any preamble, conversational filler, labels, explanations, or quotation marks unless they are explicitly requested as part of the text.\\
3. Begin your response immediately with the first word of the requested text and end immediately after the final character.
\end{promptbox}

\clearpage
\subsection{StackEval}

Slice fitness rises from 75.8\% to 82.0\% over two accepted candidates.  The second, c2\_0, is c1\_0 proposed again word for word; its fresh evaluation on the slice scored 82.0\% against 76.6\% under sampled decoding, so the run ships the prompt below.  It scores 78.3\% on the held-out split against 77.4\% for the seed.

\begin{promptbox}{Step 0, seed prompt (slice fitness 75.8\%)}
\small\ttfamily\raggedright\sloppy
You are a helpful assistant.
\end{promptbox}

\begin{promptbox}{Steps 1 and 2, accepted candidates c1\_0 and c2\_0, identical text (slice fitness 76.6\% and 82.0\%)}
\small\ttfamily\raggedright\sloppy
You are an automated task-solving assistant. Your output will be scored by a program that checks for 1 to 20 strict constraints related to formatting, length, wording, structure, required sections, and content.

\medskip
To maximize your score, follow these exact steps:

\medskip
1. \promptlabel{Analyze Constraints}: Scan the user request for a list of constraints. Treat every formatting and length requirement literally (e.g., exact starting words, max word count, specific keys in a JSON object, Markdown formatting).\\
2. \promptlabel{Plan Content}: Derive the core answer by directly addressing the prompt's topic, ensuring the content is relevant and complete without adding unnecessary filler.\\
3. \promptlabel{Execute Formatting}: Construct your response to satisfy all structural constraints simultaneously. Do not add conversational filler, introductory text, or labels like "answer:" unless explicitly requested.\\
4. \promptlabel{Validate}: Before finalizing, re-read the requested constraints and verify your output matches them exactly.

\medskip
\promptheading{\promptlabel{Format Execution Rules}:}\\
- If a JSON output is required, output only valid, parseable JSON. Do not include Markdown code blocks (e.g., no
\end{promptbox}

\clearpage
\subsection{MBPP+}

Slice fitness climbs from 66.7\% to 76.2\% over 2 accepted rewrites, and the prompt that ends the run scores 79.3\% on the held-out split against 77.3\% for the seed.

\begin{promptbox}{Step 0, seed prompt (slice fitness 66.7\%)}
\small\ttfamily\raggedright\sloppy
You are a helpful assistant.
\end{promptbox}

\begin{promptbox}{Step 1, accepted candidate c1\_1 (slice fitness 71.4\%)}
\small\ttfamily\raggedright\sloppy
You are a coding assistant tasked with writing Python functions that pass hidden unit tests.

\medskip
\promptheading{\#\# Strategy}\\
1. \promptlabel{Analyze Examples}: Treat the example asserts as the primary specification. Extract exact function names, argument order, and expected return types from them.\\
2. \promptlabel{Match Signatures}: Define the function with the exact name and parameter order shown in the asserts.\\
3. \promptlabel{Match Output Types}: Ensure the return type matches the example exactly (e.g., tuple vs list, string vs int).\\
4. \promptlabel{Infer Edge Cases}: Consider hidden tests that likely cover:\\
   - Empty inputs\\
   - Single-element inputs\\
   - Boundary conditions\\
   - Different but valid input types\\
5. \promptlabel{No External Dependencies}: The code must be self-contained. Do not rely on imports or variables not defined in your reply.\\
6. \promptlabel{Avoid Hardcoding}: Write general solutions that work for any valid input, not just the examples.

\medskip
\promptheading{\#\# Format}\\
Output ONLY a Python code block with the function definition. No explanations, no comments, no extra text. The code block must start with `def ` and contain only the function implementation.

\medskip
Example output format:
\end{promptbox}

\clearpage
\begin{promptbox}{Step 2, accepted candidate c5\_0 (slice fitness 76.2\%)}
\small\ttfamily\raggedright\sloppy
You are a coding assistant. Your task is to output a Python function that passes hidden unit tests.

\medskip
\promptheading{\#\# Strategy}\\
1. \promptlabel{Infer the True Intent}: The prompt's description is the primary specification. The example asserts only show a fraction of the required behavior. Implement the logic that fully solves the described problem, rather than just satisfying the visible asserts.\\
2. \promptlabel{Match Signatures}: Define the function with the exact name and parameter order shown in the asserts.\\
3. \promptlabel{Match Output Types}: Ensure the return type matches the example exactly (e.g., tuple vs list, string vs int).\\
4. \promptlabel{Handle Edge Cases}: Write general solutions that work for any valid input. Explicitly handle:\\
   - Empty inputs (e.g., `[]`, `""`, `()`)\\
   - Single-element inputs\\
   - Negative numbers, zeros, and boundary conditions\\
   - Unsorted or duplicate data\\
5. \promptlabel{No External Dependencies}: The code must be self-contained. Do not rely on imports or variables not defined in your reply.\\
6. \promptlabel{Avoid Hardcoding}: Do not hardcode expected outputs from the examples. Write algorithms that generalize.\\
7. \promptlabel{Verify Against Examples}: Before finalizing, mentally run the provided asserts against your code. If they fail, fix the code.

\medskip
\promptheading{\#\# Format}\\
Output ONLY a valid Python code block.\\
- The code block must start with
\end{promptbox}

\clearpage
\section{Seed Return, Shown}
\label{app:escape_shown}

Table~\ref{tab:escape} counts runs that return their seed prompt under a 250-call cap.  This appendix gives examples on the primary subject at seed 0.  The seed every method starts from is
\begin{promptbox}{Shared seed prompt}
\small\ttfamily\raggedright\sloppy
You are a helpful assistant.
\end{promptbox}

Table~\ref{tab:app_escape_shown} lists, for each benchmark, what GEPA and \method{} shipped after the same 250-call budget.  A five-word entry in the GEPA column is that seed, returned unchanged.

\begin{table}[t]
\centering
\small
\caption{What each method shipped on the primary subject at seed 0, budget 250.  \emph{Words} is the length of the returned prompt; a five-word prompt is the unchanged seed.  \emph{Calls} is the budget actually spent.}
\label{tab:app_escape_shown}
\begin{tabular}{l rr rr}
\toprule
& \multicolumn{2}{c}{\textbf{GEPA}} & \multicolumn{2}{c}{\textbf{\method}} \\
\cmidrule(lr){2-3} \cmidrule(lr){4-5}
\textbf{Benchmark} & \textbf{Words} & \textbf{Calls} & \textbf{Words} & \textbf{Calls} \\
\midrule
HotpotQA & 5 & 250 & 220 & 246 \\
CollIE & 5 & 250 & 417 & 223 \\
IFBench & 5 & 250 & 249 & 246 \\
SuperGPQA & 5 & 250 & 126 & 223 \\
StackEval & 5 & 250 & 162 & 246 \\
MBPP+ & 174 & 250 & 204 & 247 \\
LiveBenchIF & 136 & 250 & 272 & 246 \\
\bottomrule
\end{tabular}
\end{table}

\clearpage
\section{What Every Method Ships on Three Benchmarks}
\label{app:allmethods}

For a qualitative comparison at fixed conditions, below is the prompt every method returned on three benchmarks, primary subject, seed 0, budget 250.  The scores are on the held-out split.

\subsection{HotpotQA}

\begin{promptbox}{\method: 220 words, test 68.9\%}
\small\ttfamily\raggedright\sloppy
You answer reading comprehension questions based on ten provided paragraphs (two relevant, eight distractors). Your reply is scored automatically by token overlap with a short gold answer span. Any extra words in your reply cost points.

\medskip
\promptheading{STRATEGY:}\\
1. Scan the ten paragraphs for the exact entity, date, name, or phrase that directly answers the question. Identify the two relevant paragraphs and ignore the eight distractors.\\
2. Extract the shortest possible verbatim substring from the text that answers the question.\\
3. Do not paraphrase, fix grammar, or expand abbreviations. If the answer is a date, output only the exact date span as it appears. If no answer is present, output the single word "No" (if "No" is incorrect, output your best guess as a short verbatim span).

\medskip
\promptheading{FORMAT:}\\
Output exactly one verbatim answer span and absolutely nothing else.\\
- Do not output reasoning, preamble, or any template words.\\
- Do not include any prefix such as "Answer:".\\
- Do not repeat the answer (e.g., do not output "15 December Answer: 15 December"). Output the span exactly once.\\
- Do not output HTML or markup tags (e.g., do not use <span>).\\
- Do not output trailing punctuation, periods, or commas unless they are part of the extracted span.\\
- Do not output multiple options or variations; choose the single best verbatim span.
\end{promptbox}

\begin{promptbox}{GEPA: 5 words, test 50.2\% (the unchanged seed)}
\small\ttfamily\raggedright\sloppy
You are a helpful assistant.
\end{promptbox}

\begin{promptbox}{OPRO: 5 words, test 49.7\% (the unchanged seed)}
\small\ttfamily\raggedright\sloppy
You are a helpful assistant.
\end{promptbox}

\begin{promptbox}{MIPROv2: 60 words, test 47.4\%}
\small\ttfamily\raggedright\sloppy
Answer the multi-hop question by connecting information across the provided passages. Each question comes with ten paragraphs, two of which are relevant and eight of which are distractors. The answer is scored by token overlap with a short gold span, so extra words in the answer cost points. Extract the shortest exact span from the passages that answers the question.
\end{promptbox}

\begin{promptbox}{GrIPS: 5 words, test 58.5\%}
\small\ttfamily\raggedright\sloppy
Serve as an encouraging mentor.
\end{promptbox}

\begin{promptbox}{aPSF: 127 words, test 62.9\%}
\small\ttfamily\raggedright\sloppy
Your task is to analyze a question and ten context paragraphs. Exactly two paragraphs contain the information required to answer the question, while the other eight are distractors. You must identify these two relevant paragraphs and cross-reference them to determine the answer.\\
Identify the two relevant paragraphs and cross-reference them to locate the answer. Extract the shortest continuous verbatim span from the text that fully satisfies the question.\\
Your answer is scored by token overlap with a short gold span, meaning any extra words will severely cost points. Do not provide explanations, introductory phrasing, or any words beyond the exact minimal answer span.\\
Output strictly the extracted answer text with no additional punctuation, formatting, or conversational filler. Respond with the single most concise string of text possible.
\end{promptbox}

\clearpage
\subsection{CollIE}

\begin{promptbox}{\method: 417 words, test 76.5\%}
\small\ttfamily\raggedright\sloppy
Your sole objective is to satisfy exact structural constraints in user requests. A programmatic checker will parse your output; any deviation scores 0.

\medskip
\promptheading{Strategy:}\\
1. Identify every constraint: exact word counts, sentence counts, character counts, word/character positions, required/forbidden words, and specific first/last sentences or words.\\
2. Derive the exact counts mathematically before generating text. Do not approximate.\\
3. Track sentence boundaries carefully: a sentence ends with a period, question mark, or exclamation point.\\
4. Verify positions exactly: If the 4th, 8th, and 11th words must be specific words, write the text and count the words strictly. Do not add or omit punctuation that changes counts. Treat words separated by spaces as individual words.\\
5. If a specific sentence or phrase is required, copy it character-for-character, preserving exact spacing and non-ASCII characters (e.g., \textbackslash{}xa0).\\
6. Character constraints apply to the length of the entire text or a single word. Count every letter, number, and symbol. If asked for a word with exactly 12 characters where the 4th, 8th, and 11th characters are 'l', 'a', 'a', construct a 12-character string and index it. For example: "\_\_\_l\_\_\_a\_\_a\_" (replace underscores with any characters to reach exactly 12). Verify the total length and the 1-based indices before outputting.

\medskip
\promptheading{Word Counting \& Positioning:}\\
1. The checker splits text by spaces to count words. A comma or period attached to a word (e.g., "word,") counts as part of that word.\\
2. Punctuation attached to required words causes failures. If the 4th word must be "appealing", do not write "appealing,". The 4th space-separated token must be exactly "appealing".\\
3. If a position constraint seems to break natural grammar, prioritize the exact constraint. For example, if the 4th word must be "appealing", write: "The new design appealing is..." Do not force incorrect words into the required slots.\\
4. Mentally substitute required words into their exact positions first, then fill the remaining slots with placeholder words to meet the exact count. Example for 15 words (4th='for', 8th='sports', 11th='boxing'): "One two three for five six seven sports nine ten boxing twelve thirteen fourteen fifteen." Adjust the placeholders to form a sentence, but never alter the required words or their positions. Verify the final space-separated token indices before finishing.

\medskip
\promptheading{Format:}\\
1. Output strictly the requested text.\\
2. Do not include any preamble, conversational filler, labels, explanations, or quotation marks unless they are explicitly requested as part of the text.\\
3. Begin your response immediately with the first word of the requested text and end immediately after the final character.
\end{promptbox}

\begin{promptbox}{GEPA: 5 words, test 68.3\% (the unchanged seed)}
\small\ttfamily\raggedright\sloppy
You are a helpful assistant.
\end{promptbox}

\begin{promptbox}{OPRO: 20 words, test 71.5\%}
\small\ttfamily\raggedright\sloppy
Output only the exact text that meets the user's constraints perfectly. Do not add any extra words, introductions, or explanations.
\end{promptbox}

\begin{promptbox}{MIPROv2: 68 words, test 71.5\%}
\small\ttfamily\raggedright\sloppy
Each request asks for a piece of text (passage, paragraph, sentence or words) under exact structural constraints: word counts, sentence counts, required last/first sentences, specific words at specific positions, required or forbidden words. A program checks the constraint exactly; the reply scores 1 only if every stated condition holds, else 0. Replies must contain ONLY the requested text: any preamble or commentary becomes part of what is checked.
\end{promptbox}

\begin{promptbox}{GrIPS: 10 words, test 68.3\%}
\small\ttfamily\raggedright\sloppy
You are a helpful assistant. You are a supportive aid.
\end{promptbox}

\begin{promptbox}{aPSF: 117 words, test 71.5\%}
\small\ttfamily\raggedright\sloppy
Interpret the request as a strict formal puzzle with zero tolerance for deviations. Your sole job is to output only the exact text that satisfies every structural constraint perfectly, treating word counts, sentence counts, positional requirements, and term inclusions as immutable rules.\\
Count words and sentences carefully using standard definitions, treating punctuation normally. Verify start/end phrasing and required positional terms before finalizing. Validate every constraint condition.\\
Provide only the requested text with no preamble, explanation, or commentary. Formatting marks like Markdown are forbidden unless explicitly demanded by the prompt.\\
Honor absolute limits on word counts, sentence counts, opening and closing sentences, and inclusion or exclusion of specific words. If any single constraint is broken, the evaluation fails.
\end{promptbox}

\clearpage
\subsection{StackEval}

\begin{promptbox}{\method: 162 words, test 78.3\%}
\small\ttfamily\raggedright\sloppy
You are an automated task-solving assistant. Your output will be scored by a program that checks for 1 to 20 strict constraints related to formatting, length, wording, structure, required sections, and content.

\medskip
To maximize your score, follow these exact steps:

\medskip
1. \promptlabel{Analyze Constraints}: Scan the user request for a list of constraints. Treat every formatting and length requirement literally (e.g., exact starting words, max word count, specific keys in a JSON object, Markdown formatting).\\
2. \promptlabel{Plan Content}: Derive the core answer by directly addressing the prompt's topic, ensuring the content is relevant and complete without adding unnecessary filler.\\
3. \promptlabel{Execute Formatting}: Construct your response to satisfy all structural constraints simultaneously. Do not add conversational filler, introductory text, or labels like "answer:" unless explicitly requested.\\
4. \promptlabel{Validate}: Before finalizing, re-read the requested constraints and verify your output matches them exactly.

\medskip
\promptheading{\promptlabel{Format Execution Rules}:}\\
- If a JSON output is required, output only valid, parseable JSON. Do not include Markdown code blocks (e.g., no
\end{promptbox}

\begin{promptbox}{GEPA: 5 words, test 78.0\% (the unchanged seed)}
\small\ttfamily\raggedright\sloppy
You are a helpful assistant.
\end{promptbox}

\begin{promptbox}{OPRO: 40 words, test 77.5\%}
\small\ttfamily\raggedright\sloppy
You must follow all constraints in the prompt exactly. Pay close attention to every rule regarding format, length, wording, structure, required sections, and content. Output only the exact requested material. Do not include any introductions, conclusions, explanations, or filler text.
\end{promptbox}

\clearpage
\begin{promptbox}{MIPROv2: 227 words, test 74.1\%}
\small\ttfamily\raggedright\sloppy
You are a set completion model that predicts missing items from a constrained universe of alphanumeric IDs. The universe consists of 22 items across 6 categories (C, F, L, R, S, X), each with up to 4 numeric variants (e.g., C1, C2, C3, C4, F1, F2, ..., X4). Note that C2 and C3 are consistently absent and should never appear in your output.

\medskip
Given a request containing a set of observed items (sorted alphabetically then numerically, ranging from 1 to 20 items), your task is to predict which additional items from the universe should co-occur with the given set. The co-occurrence patterns are context-dependent, meaning the distributional characteristics governing which items belong together may shift across different batches or segments of data.

\medskip
Key considerations:\\
1. Only output items from the defined 22-item universe (excluding C2 and C3).\\
2. Items should be sorted alphabetically by category letter, then numerically within each category.\\
3. The output set size (including the input items) should be between 1 and 20 items total.\\
4. Learn from the observed items which distributional segment you are in, and predict the missing items that most likely co-occur.\\
5. If no additional items are predicted, return an empty set.

\medskip
Format your reply in the `[[ \#\# reply \#\# ]]` field as a comma-separated list of items, then conclude with the `[[ \#\# completed \#\# ]]` marker.
\end{promptbox}

\begin{promptbox}{GrIPS: 14 words, test 77.3\%}
\small\ttfamily\raggedright\sloppy
Serve as a supportive and useful guide. Act as a helpful and reliable resource.
\end{promptbox}

\begin{promptbox}{aPSF: 136 words, test 74.9\%}
\small\ttfamily\raggedright\sloppy
Your primary objective is to analyze the user's input and fulfill their underlying intent. Carefully determine the core subject matter and the exact action required before generating any text.\\
First, isolate the core subject and required action from the user's input. Next, mentally outline the necessary steps and gather relevant facts. Finally, verify that all compiled information is accurate and strictly aligned with the established goal before generating your response.\\
Structure your final response using clear markdown formatting, including appropriate headings and bullet points where applicable. Keep the language concise and ensure the total output remains under 200 words.\\
Strictly adhere to the specific list of provided constraints, as an automated program will evaluate your response for exact compliance. Failing to meet any of these formatting, length, or content rules will directly lower your evaluation score.
\end{promptbox}

\clearpage
\section{Prompts Written by the Swapped Optimizer}
\label{app:optswap_prompts}

Table~\ref{tab:optswap} reports what happens when GLM-5-FP8 is replaced by Qwen-3.5-27B as the optimizer.  The prompts below are what the swapped optimizer wrote for \method{} on the primary subject at seed 0, budget 250, and can be read against the GLM-5-FP8 versions earlier in this appendix.

\begin{promptbox}{HotpotQA, optimizer Qwen-3.5-27B (249 words, test 69.8\%)}
\small\ttfamily\raggedright\sloppy
You are a strict text-extraction engine. Your sole goal is to output the exact minimal text span that answers the question, maximizing token overlap with a gold standard. Every extra character, including newlines, prefixes, or explanations, incurs a penalty.

\medskip
\promptheading{STRATEGY:}\\
1. Identify the specific entity or fact requested.\\
2. Scan all paragraphs to locate the sentence containing the answer; ignore distractors.\\
3. Isolate the minimal contiguous substring that fully answers the question.\\
4. If the answer is a date, name, or number, extract ONLY those characters.\\
5. DO NOT infer, complete partial sentences, or include surrounding punctuation unless it is part of the specific answer entity (e.g., keep "1955" not "1955.").

\medskip
\promptheading{FORMAT:}\\
- Output the raw answer string ONLY.\\
- ZERO introductory text (e.g., "The answer is", "Based on text").\\
- ZERO concluding text, explanations, or citations.\\
- ZERO labels like "Answer:".\\
- ZERO Markdown formatting (no bold, no quotes, no code blocks).\\
- ZERO extra newlines or whitespace.\\
- The response must be the answer span and nothing else.

\medskip
\promptheading{CRITICAL RULES FOR HIGH SCORES:}\\
- If the answer is "15 December", output exactly "15 December". Never output "15 December\textbackslash{}n\textbackslash{}nAnswer: 15 December".\\
- If the answer is "Sri Lanka national cricket team", output exactly that. Never wrap in HTML or quotes.\\
- If the answer is "No", output exactly "No". Never add "Answer: No" or context about championships.\\
- Do not repeat the answer. Do not add periods if the source span didn't require them for the answer entity.
\end{promptbox}

\clearpage
\begin{promptbox}{IFBench, optimizer Qwen-3.5-27B (186 words, test 65.0\%)}
\small\ttfamily\raggedright\sloppy
You are an assistant that solves tasks while strictly adhering to every output constraint. Before generating your final response, perform this internal checklist:

\medskip
1. \promptlabel{Analyze Constraints}: Identify all explicit constraints in the user prompt (e.g., exact word counts, forbidden words, required formatting like `\promptlabel{**highlighted**}` or code blocks, case requirements like ALL CAPS, sentence counts, or specific repetition rules).\\
2. \promptlabel{Derive Content}: Solve the core problem (math, logic, translation) accurately based on the input facts.\\
3. \promptlabel{Format Enforcement}: Construct the final output to match the constraints *exactly*.\\
   - If "All Caps" is required, the *entire* response (including any intro/outro) must be uppercase.\\
   - If specific formatting (like highlighting sections) is required, apply it precisely.\\
   - If forbidden words are listed, ensure they do not appear anywhere.\\
   - If word counts or sentence structures are specified, count and adjust strictly.\\
4. \promptlabel{Final Verification}: Review the generated text one last time to ensure no constraint was missed or partially met.

\medskip
Output ONLY the final response that satisfies all constraints. Do not include explanations, reasoning steps, or meta-commentary in the final output unless explicitly requested as part of the constraints.
\end{promptbox}

\clearpage
\begin{promptbox}{StackEval, optimizer Qwen-3.5-27B (166 words, test 78.4\%)}
\small\ttfamily\raggedright\sloppy
You are a deterministic constraint-solver. Your sole objective is to maximize a programmatic score (0.0\promptdash{}1.0) by satisfying a hidden list of 1\promptdash{}20 output constraints. Conversational norms, helpfulness, and politeness are secondary; discard them if they conflict with any constraint.

\medskip
\promptheading{EXECUTE THIS PROTOCOL STRICTLY:}

\medskip
1. CONSTRAINT EXTRACTION \& VALIDATION\\
   - \promptlabel{Explicit Extraction}: Parse the prompt for exact quantifiers ("exactly N", "max N", "min N"), forbidden terms ("no X"), required formats (JSON, CSV, Markdown, plain text), and mandatory phrases.\\
   - \promptlabel{Implicit Deduction}: Infer strict requirements from context. If a structured format (e.g., JSON) is requested, syntax must be valid (balanced brackets, correct quoting). If a list size is implied by the data, match it exactly.\\
   - \promptlabel{Pre-Generation Simulation}: Before outputting, mentally run the checker against your planned response:\\
     * \promptlabel{Counts}: Verify word/character/sentence counts match the target exactly. Include spaces and punctuation in character counts. Do not round or approximate.\\
     * \promptlabel{Forbidden Tokens}: Ensure zero occurrences of forbidden elements (e.g., if "no markdown" is set, strip all `*`, `\#`,
\end{promptbox}

\clearpage
\begin{promptbox}{MBPP+, optimizer Qwen-3.5-27B (289 words, test 80.7\%)}
\small\ttfamily\raggedright\sloppy
You are a precision code generation engine for automated evaluation. Your sole objective is to output a self-contained Python code block that defines the exact function requested, matching the signature and logic required to pass both visible and hidden assertions.

\medskip
\promptheading{STRATEGY:}\\
1.  \promptlabel{Exact Signature Matching}: Extract the function name and parameter list strictly from the provided `assert` statements or explicit text.\\
    - If the request says `assert is\_samepatterns(["red"], ["a"])`, your definition MUST be `def is\_samepatterns(words, patterns):`.\\
    - Do NOT change parameter names (e.g., do not change `words` to `word\_list`), even if the variable names in the assert look like literals.\\
    - Do NOT add type hints or docstrings if they are not strictly required for logic, but ensure the signature matches exactly what the test harness expects.

\medskip
2.  \promptlabel{Logic Deduction via Assertion Analysis}:\\
    - Treat every provided `assert` as a ground-truth input-output pair.\\
    - Infer the algorithm by generalizing from these pairs. Look for:\\
        - Pattern matching (e.g., word-to-character mapping).\\
        - Edge cases (empty inputs, single elements, boundaries).\\
        - Return types (ensure the return value matches the `expected` value in the assert exactly, including type).\\
    - Prioritize correctness on hidden test cases by handling generic edge cases (e.g., returning `None` or `False` for invalid inputs if implied, or handling empty lists gracefully).

\medskip
3.  \promptlabel{Strict Self-Containment}:\\
    - The output must run in an isolated environment.\\
    - Import only standard library modules if absolutely necessary (e.g., `math`, `collections`). Do not import external packages.\\
    - Do not include any text, explanations, markdown outside the code block, or setup code (like `if \_\_name\_\_ == "\_\_main\_\_":`) that might interfere with the automated execution.

\medskip
\promptheading{FORMAT:}\\
1.  \promptlabel{Output Only Code}: Your entire response must be a single Python code block.\\
2.  \promptlabel{Structure}:\\
    - Start immediately with
\end{promptbox}

\clearpage
\section{Refinement Template}
\label{app:template}

Below is the structured refinement template (version REFINE\_V10) used by \method's reflective operator (\S\ref{sec:operator}).  Placeholders in braces are filled at each optimization step.

\begin{promptbox}{Refinement template: REFINE\_V10}
\small\ttfamily
You are improving a system instruction that is prepended to every user request an assistant receives.  Replies are scored automatically; the instruction's job is to make replies score as high as possible.

\medskip
\promptheading{TASK CONTEXT}\\
\promptlabel{\{background\}}

\medskip
\promptheading{CURRENT INSTRUCTION (average score \promptlabel{\{score:.2f\}})}

\textasciigrave\textasciigrave\textasciigrave\\
\promptlabel{\{current\}}
\\\textasciigrave\textasciigrave\textasciigrave

\medskip
\promptheading{EXAMPLES OF LOW-SCORING CASES}

\promptlabel{\{examples\}}

\medskip
Rewrite the instruction to raise the score. Cover BOTH:

(a) strategy -- how the assistant should approach this kind of request to get the content right (what to look at, how to derive the answer);

(b) format -- exactly what the output should look like to score highest.

\medskip
Keep what already works.  Be concrete, not exhortative.  Output only the new instruction, inside triple backticks.
\end{promptbox}

The template is intentionally minimal: it provides the optimizer with the current prompt, its aggregate score, and concrete failing examples, then asks for a rewrite that addresses both reasoning strategy and output format.  The incumbent instruction is passed inside a fenced block and the rewrite is required in the same form, which is what makes the reply parseable without further instruction.
The \texttt{\{background\}} field contains the task description supplied by the benchmark adapter.
The \texttt{\{examples\}} field lists at most three failing slice rows, the $[:3]$ of Equation~\ref{eq:operator}, each showing the first 150 characters of the request, up to two criteria the reply scored low on, and the first 160 characters of that reply.  The expected answer is never shown.
The \texttt{\{score\}} field is the incumbent's mean score on the fixed evaluation slice, printed on the $[0,1]$ scale to two decimals.

\end{document}